\pdfoutput=1

\documentclass[11pt]{article}

\usepackage[]{acl}

\usepackage{times}
\usepackage{latexsym}

\usepackage[T1]{fontenc}

\usepackage[utf8]{inputenc}

\usepackage{microtype}

\usepackage{inconsolata}

\usepackage{graphicx}
\usepackage{multirow}

\title{Towards Detecting, Recognizing, and Parsing the Address Information from Bangla Signboard: A Deep Learning-based Approach}

\author{Hasan Murad, Mohammed Eunus Ali\\
Department of Computer Science and Engineering\\
Bangladesh University of Engineering and Technology, Bangladesh\\
\texttt hasanmurad@cuet.ac.bd, \texttt eunus@cse.buet.ac.bd}

\begin{document}
\maketitle
\begin{abstract}
Retrieving textual information from natural scene images is an active research area in the field of computer vision with numerous practical applications. Detecting text regions and extracting text from signboards is a challenging problem due to special characteristics like reflecting lights, uneven illumination, or shadows found in real-life natural scene images. With the advent of deep learning-based methods, different sophisticated techniques have been proposed for text detection and text recognition from the natural scene. Though a significant amount of effort has been devoted to extracting natural scene text for resourceful languages like English, little has been done for low-resource languages like Bangla. In this research work, we have proposed an end-to-end system with deep learning-based models for efficiently detecting, recognizing, correcting, and parsing address information from Bangla signboards. We have created manually annotated datasets and synthetic datasets to train signboard detection, address text detection, address text recognition, address text correction, and address text parser models. We have conducted a comparative study among different CTC-based and Encoder-Decoder model architectures for Bangla address text recognition. Moreover, we have designed a novel address text correction model using a sequence-to-sequence transformer-based network to improve the performance of Bangla address text recognition model by post-correction. Finally, we have developed a Bangla address text parser using the state-of-the-art transformer-based pre-trained language model.
\end{abstract}

\section{Introduction}\label{intro}
Text is the written form of human language and is considered one of the oldest and most powerful creations of human civilization. Text is the most effective and reliable means of communication, collaboration, and documentation. Our surroundings are rich with textual information that assists in interpreting the world around us. Automatic natural scene text detection and recognition have become an active research problem due to numerous real-life practical applications \cite{lin2020review, park2010automatic, wojna2017attention, yi2012assistive}. A common source of natural scene text is the signboard which contains crucial information such as the name of the business organization and its current address. However, retrieving such information automatically from the signboard is a challenging problem due to the complex background with interference, high variation and diversity in natural scene text, and imperfect lighting during imaging \cite{park2010automatic}. With the introduction of sophisticated deep learning-based techniques, we are witnessing a dramatic improvement in the field of computer vision and natural language processing (NLP). In this research work, we aim to develop deep learning-based models for efficiently detecting, recognizing, and parsing address information from Bangla signboard.

 With the recent advent in the field of computer vision using deep learning-based techniques, natural scene text detection and recognition have become an active research problem due to numerous real-life practical applications \cite{lin2020review, park2010automatic, wojna2017attention, yi2012assistive}. 
 

 Business organization manually annotates on mapping platforms like Google Maps\footnote{https://www.google.com/maps/} or OpenStreetMap\footnote{https://www.openstreetmap.org/} so that the customer easily locate these organizations on mapping platforms. However, a significant number of business organizations in developing countries like Bangladesh remain unannotated in mapping platforms due to a lack of technical knowledge of the business owner. A popular service of Google is Google Street View\footnote{https://www.google.com/streetview/} which provides interactively panoramas along the street of the metropolitan area. From the street imagery, the signboard of the different business organizations can be detected. An effective text detector and extractor from signboards facilitate automatic annotation of mapping platforms from street imagery \cite{wojna2017attention}. 

 
  According to a recent report by the World Health Organization (WHO), around 2.2 billion people are suffering from near or distance vision impairment where 1 billion people are facing moderate to severe blindness\footnote{https://www.who.int/news-room/fact-sheets/detail/blindness-and-visual-impairment}. Different AI-driven technologies have been introduced to alleviate the day-to-day problems faced by visually impaired people \cite{kuriakose2022tools}. Developing assistive navigation applications is an active research area among practitioners \cite{khan2021analysis}.   Extracting text information from signboards facilitates to development of camera-based navigation applications for visually impaired people \cite{yi2012assistive}. We can develop such a navigation application for smartphones or embedded devices such as smart glass.


 Mapping platform users usually search for a location or organization by providing a raw input address text which needs to be efficiently processed to find the point of interest on the map. The raw input address text needs to be parsed to find out the segment of the address to effectively search on a mapping platform. Only rule-based techniques fail to parse all the variations of the user address input. Therefore, automatic parsing of raw input address text using natural language processing facilitates efficient searching on map applications. 

 
 During the registration on an information system, a user provides their present or permanent address in a raw text field. However, before inserting the address text into the database, the address text needs to be parsed to find different address components. An address parsing using natural language processing facilitates automatic insertion on a database from raw input address text \cite{mokhtari2019tagging}.

\begin{figure}[!h]
  \centering
  \includegraphics[width=0.45\textwidth]{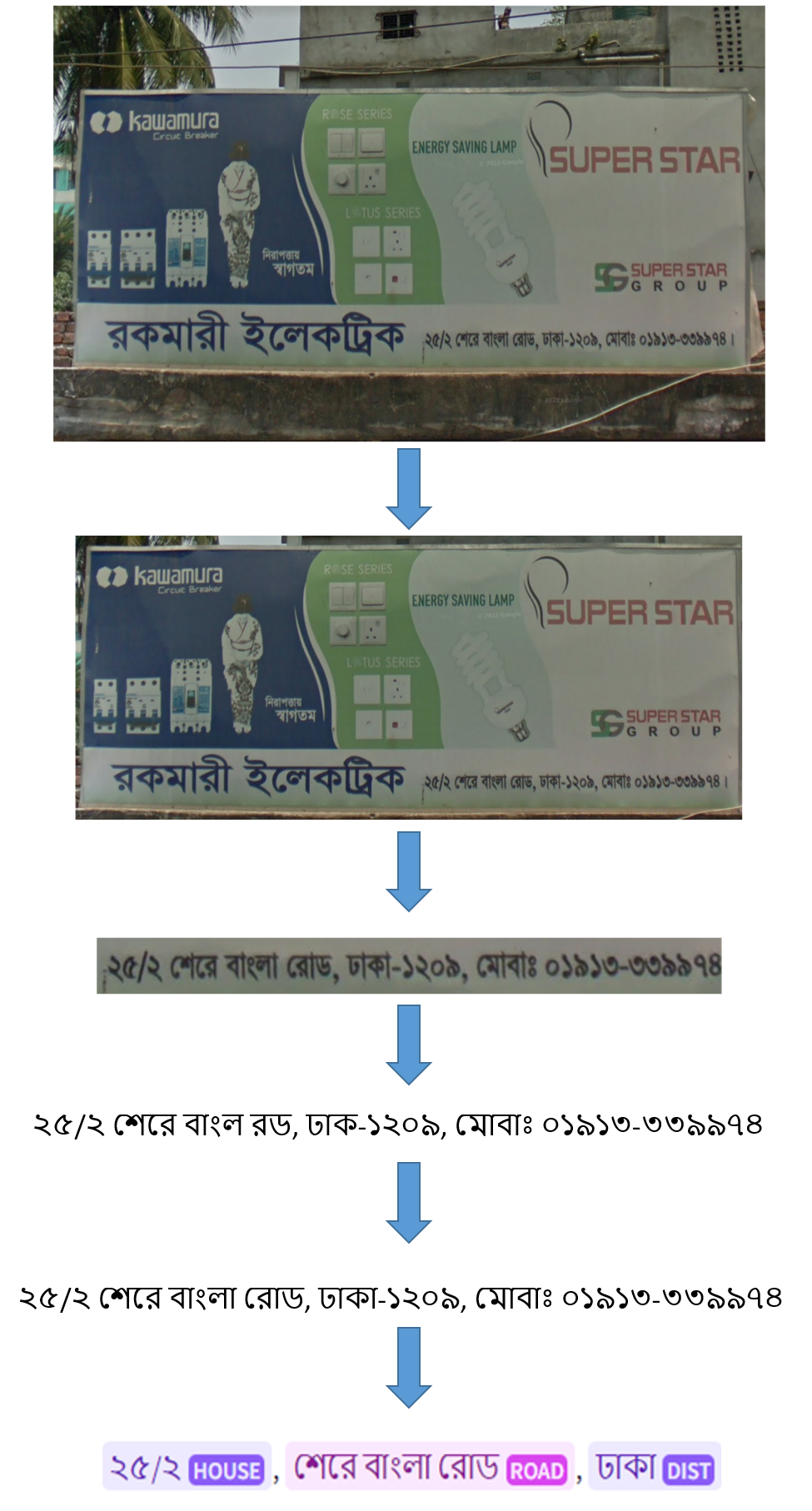}
  \caption{An overview of the research problem}
  \label{fig:problem statement}
\end{figure}

 Figure \ref{fig:problem statement} shows an overview of our research problem. We develop an end-to-end system for detecting, recognizing, and parsing the address information from the signboard. First, we have to detect the signboard from the raw natural scene image. The next step is to detect the address text region from the signboard.  After detecting the address text region, the address text is recognized from the cropped address text. To minimize the recognition error, we need to conduct necessary post correction on the recognized address text. Finally, the corrected address text is parsed to identify the different segments of the address.
 
 With the advent of deep learning-based methods, different sophisticated techniques have been proposed for text detection and text recognition from the natural scene \cite{long2021scene, khan2021deep, chen2021text}. The Yolo-based model is the state-of-the-art object detection model for text detection \cite{haifeng2020natural}. In literature, different segmentation-based \cite{ahmed2019complete, isthiaq2020ocr} and segmentation-free techniques \cite{shi2016end, he2016reading, liu2018squeezedtext} have been deployed for text recognition from images. Due to the challenges involved in natural scene text, segmentation-free techniques like Connectionist Temporal Classification (CTC) based \cite{shi2016end, he2016reading} or Encoder-Decoder \cite{liu2018squeezedtext} models outperform segmentation-based techniques for text recognition. However, segmentation-free techniques for text recognition remain nearly unexplored for low-resource languages like Bangla. Moreover, no correction model has been found to improve the performance of the text recognition model by post correction. Different deep learning-based sequence-to-sequence models \cite{mokhtari2019tagging, abid2018deepparse, yassine2021leveraging} with RNN, LSTM, and Bi-LSTM units have been proposed for address parsing. However, the state-of-the-art transformer-based pre-trained models have not been explored for the address parsing problem. Though a significant amount of effort has been devoted to extracting address information from signboards for resourceful languages like English \cite{wojna2017attention}, Korean \cite{park2010automatic}, or Japanese \cite{nomura2014automatic}, little has been done for low-resource languages like Bangla.  

In this research work, we develop deep learning-based models for efficiently detecting, recognizing, and parsing address information from Bangla signboards.  The main objectives of our study are as follows:
\begin{itemize} 
\item To conduct an in-depth analysis of different segmentation-free CTC-based and Encoder-Decoder model architectures for Bangla address text recognition. 
\item To design a novel address text correction model using a sequence-to-sequence transformer network to improve the performance of Bangla address text recognition by post-correction.
\item To develop a Bangla address text parser using the state-of-the-art transformer-based pre-trained language model.
\end{itemize}

\begin{figure}[!h]
  \centering
  \includegraphics[width=0.35\textwidth]{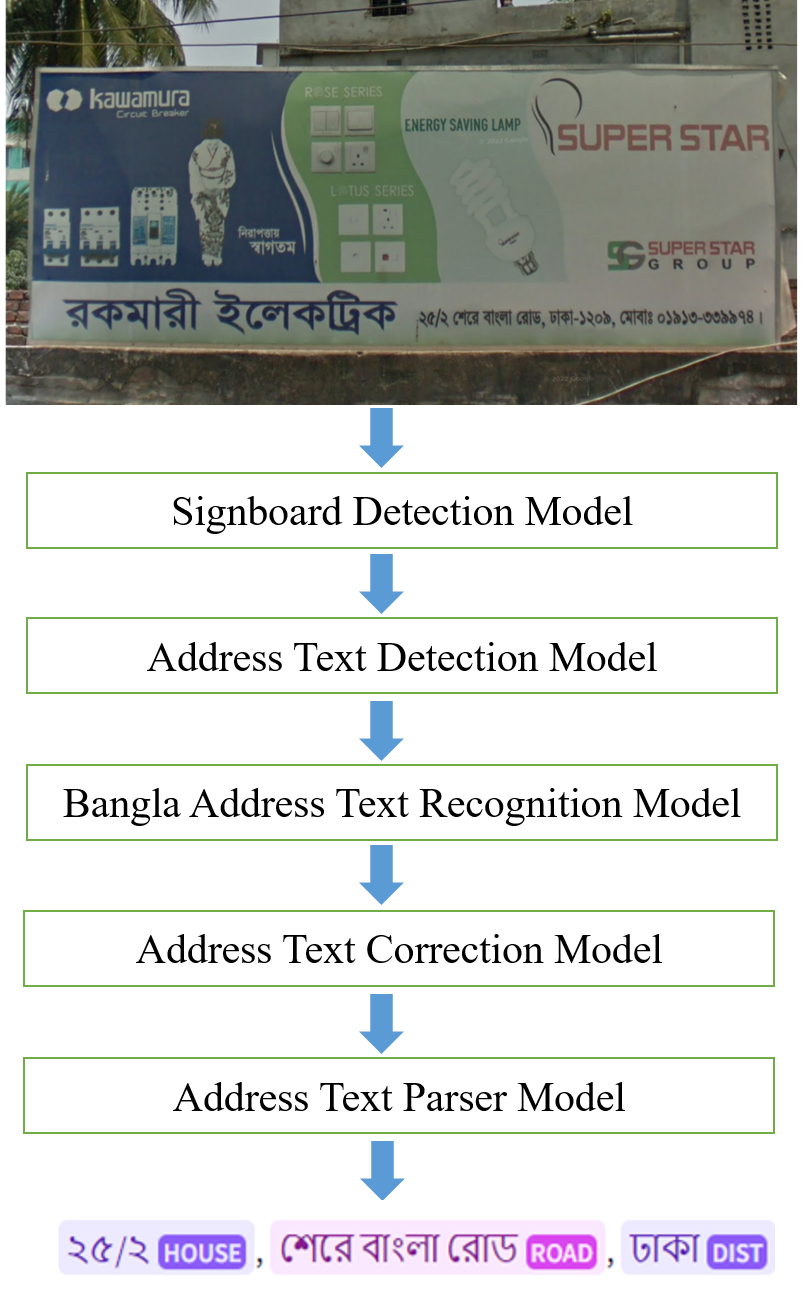}
  \caption{An overview of the proposed solution}
  \label{fig:soluton overview}
\end{figure}

To develop an end-to-end system for extracting and parsing address information from the signboard, we divide the whole system into different sub-problems:  a signboard detection model for detecting signboard from the natural scene image; an address text detection model for detecting address region from a signboard; a custom Bangla address text recognition model to extract address text from the cropped address region; an address text correction model to improve the output of the address text recognition model by post-correction; and finally an address text parser model to classify each field of an address. Figure \ref{fig:soluton overview} shows an overview of our proposed end-to-end system. 

We develop two detection models - the signboard detection model and the address text detection model. The signboard detection model is responsible for identifying the signboard region within a natural scene image. Subsequently, the address text detection model utilizes the cropped signboard image to specifically identify the portion containing the address text. We collect a large number of natural scene images containing Bangla signboards and annotate them appropriately to create the Bangla signboard detection dataset. Moreover, we crop the signboard from the natural scene image and develop a Bangla address detection dataset by annotating it.  In this research work, we train and evaluate different recent Yolo-based object detection models such as YOLOv3, YOLOv4, and YOLOv5 for both the signboard detection model and the address text detection model using the Bangla signboard detection dataset and Bangla address detection dataset respectively. 

After detecting the address text portion from the signboard, the next step is to recognize the address text from the cropped address text portion image. We have proposed two different frameworks for Bangla address text recognition - the CTC-based framework and the Encoder-Decoder framework. 

We have designed the CTC-based model which consists of three major components: convolutional layers for feature extraction; recurrent layers for sequence modeling; transcription layer with CTC loss for sequence prediction. Moreover, we have proposed an Encoder-Decoder model architecture consisting of three key components: a deep-stacked convolution neural network for feature extraction, an encoder network with a recurrent layer for sequence modeling, and finally a decoder network with an attention mechanism for transcription. 

We have created a synthetically generated Bangla address text recognition dataset to train and evaluate different CTC-based and Encoder-Decoder model architectures. We have conducted a comprehensive evaluation of various CTC-based and Encoder-Decoder models for Bangla address text recognition to determine the most effective model architecture. 

To improve the performance of the Bangla address text recognizer by post-correction, we propose an address text correction model using a transformer-based sequence-to-sequence model. To train and evaluate the address text correction model, we create a synthetically generated address correction dataset from the raw address corpus. 

Finally, we propose a state-of-the-art transformer-based pre-trained language model for Bangla address text parsing. We develop a novel Bangla address parsing dataset to train and evaluate the proposed Bangla address text parser model. Moreover, we train traditional sequence-to-sequence models with RNN, LSTM, and Bi-LSTM units to present a comparative analysis with a transformer-based pre-trained language model for Bangla address text parsing. 

The main contributions of our research work are as follows:
\begin{itemize}
\item We have created manually annotated datasets and synthetic datasets for detecting, extracting, correcting, and parsing address information from the natural scene.
\item We have conducted a performance analysis among different CTC-based and Encoder-Decoder models for Bangla address text recognition and found the best-performing model architecture.  
\item We have introduced a novel address text correction model using a sequence-to-sequence transformer network to improve the performance of Bangla address text recognition by post-correction.
\item We have developed a Bangla address text parser using the state-of-the-art transformer-based pre-trained language model.
\end{itemize}

\section{Related Works}\label{related-works}
In this section, we explore the previous research works. We divide our research work into different sub-problems such as text detection, text recognition, and address parsing. After conducting a rigorous literature review, we have found that there are both traditional machine learning-based and sophisticated deep learning-based approaches to solving each sub-problem. We present a complete picture by classifying the previous research works into a hierarchical taxonomy. We introduce the related works in a top-down approach. We organize and present the related works into different categories: (1) Scene text detection to detect and localize the text area in natural scene text; (2) Scene text recognition to extract the textual information from the detected text area and convert them into linguistic symbols; (3) End-to-end approach to performing both detection and recognition of scene text in a single pipeline; (4) Addressing parsing to classify different segments of the address text; and finally (5) Previous works related to information extraction from the signboard. We introduce each category by discussing the related works from different perspectives. Moreover, we present existing previous works in Bangla language under each category. 
\subsection{Scene Text Detection}
A significant amount of effort has been devoted to solving the scene text detection problem, especially for the English language text. There are different approaches found in the literature for scene text detection and localization \cite{naosekpam2022text, long2021scene, khan2021deep, lin2020review}. We divide these related existing works into four different sub-categories: (1) Scene text detection using statistical features such as SWT \cite{epshtein2010detecting, li2012scene} and MSER \cite{matas2004robust, neumann2011method}; (2) Scene text detection using traditional machine learning-based approaches such as sliding window-based \cite{wang2011end, pan2009text} and connected component-based \cite{chen2011robust, yin2013robust}; (3) Scene text detection using hybrid approaches \cite{jaderberg2014deep, khatib2015hybrid}, and scene text detection using sophisticated deep learning-based approaches such as object detection-based \cite{liao2017textboxes, liao2018textboxes++} and segmentation-based \cite{deng2018pixellink, qin2019algorithm}. 
\subsubsection{Scene Text Detection Using Statistical Feature-based Approaches}
For a very long time, scene text detection is an active research area in the field of computer vision. Early approaches apply traditional image features to detect the scene text. Text is localized from the background image by considering that the colors of the pixel will be similar for a single character and the background colors are different from the text colors \cite{zhong1995locating}. However, if we consider real-life scenarios, such a hypothesis fails to detect scene text on a complex background.  Moreover, a split-and-merge approach has been found to segment the text area by considering the text with similar font size and color as connected components \cite{lienhart1996automatic}. Such constraints based on the color component and font size make these methods highly dependent on the standard scenarios. Innovative features such as stroke width transform (SWT) \cite{epshtein2010detecting, li2012scene} and maximally stable extremal regions (MSER) \cite{matas2004robust} are utilized for statistical feature-based scene text detection. SWT calculates the distance between edge pixel pairs to obtain stroke width information, which is used for bottom-up integration of pixels with similar stroke widths, combined into connected components, grouped into letter candidates, and text lines and then finally clustered into chains based on specific criteria.  MSER is used to detect individual characters as Extremal Regions (ER) based on color similarity and computation complexity \cite{neumann2011method}. However, such approaches are challenging to apply for detecting colorful text in the real-life natural scene. To address the difficulties in detecting minor and blurry texts and symbols in natural scene images, a combination of the edge-enhanced MSER and Canny edge detector \cite{canny1986computational} has been suggested \cite{chen2011robust}. The complete procedure comprises geometric checks, pairing letters, constructing lines of text, and dividing them into individual words.

Although statistical feature-based methods for detecting scene text are effective in controlled settings, they are ineffective in more challenging real-world scenarios commonly encountered in natural environments.

\subsubsection{Scene Text Detection Using Machine Learning-based Approaches}
The conventional machine learning methods for detecting text in scenes involve using Sliding Window (SW) \cite{chen2004detecting, hanif2008cascade, koo2011text, pan2009text, pan2010hybrid} based and Connected Component Analysis (CCA) techniques \cite{chen2001text, huang2013text, yin2013robust, gomez2015object}. These methods locate candidate text regions by extracting handcrafted features and subsequently classify them to determine the actual text regions. However, the slide window-based methods apply a top-down approach, whereas the CCA-based technique works in a bottom-up strategy. 
\subsubsection*{Sliding Window-based Scene Text Detection}
In a top-down approach, sliding window-based scene text detection first slides a multi-scale window over the input image to identify the candidate regions and then applies a pre-trained classifier to determine whether the sub-window contains text or not. To distinguish between text and non-text regions, textural characteristics such as white space information, responses from different types of filters, and wavelet coefficients are utilized. Chen and Yuille consider such textural features to design weak classifiers and then apply the Adaboost algorithm to obtain a final strong classifier \cite{chen2004detecting}. Moreover, Support Vector Machine (SVM) \cite{kim2003texture} is also used to classify the raw image without extracting the characteristics as features. Hanif et al. propose two different types of weak classifiers using Linear Discriminant Classifier (LDC) and Log Likelihood-ratio Test (LRT) \cite{hanif2008cascade}. To identify the text regions, adaptive Meanshift is utilized \cite{comaniciu2002mean} and cascaded features such as mean difference and standard deviation are used to extract overleaping text segment \cite{hanif2009text, hanif2008cascade}. Extracting text lines is approached as an optimization problem that minimizes energy and accounts for potential interference between text lines. Promising outcomes have been achieved using Conditional Random Field (CRF) and Markov Random Field (MRF) based energy minimization methods \cite{pan2009text, koo2011text, pan2010hybrid}.

The sliding window method addresses the text region identification problem by using a classifier to calculate the likelihood of text utilizing a feature vector extracted from the individual local region. Identified neighboring text regions are finally combined to find text blocks. However, such a brute-force approach is usually slow due to its reliance on local decision-making, even though it is effective for text regions with distinct textural properties.

\subsubsection*{Connected Component Analysis-based Scene Text Detection}
In a bottom-up manner, connected component analysis methods \cite{chen2001text, huang2013text, yin2013robust, gomez2015object} extract components from the input image using different clustering algorithms or edge detection techniques, eliminate non-textual regions of the image by applying different heuristic approaches or pre-trained classifiers, and finally group the neighboring textual regions using geometric properties.  

By considering that the text region has a rectangular shape and horizontal alignment, Chen et al. propose a technique for text detection that involves extracting candidate text lines and using an SVM classifier to identify the actual text lines from the extracted candidates \cite{chen2001text}. By modifying SWT to incorporate the color component, Huang. et al. introduce the Stroke Feature Transform (SFT) algorithm which calculates the stroke width map to extract the candidate text regions and finally, classify the actual text regions using text covariance descriptors \cite{huang2013text}. Modifying MSRT to improve performance for images with poor quality, Yin et al. propose an algorithm for parent-children elimination \cite{yin2013robust}. The algorithm eliminates the extremal region from the MSER tree when the aspect ratio of the extremal region does not match a given range. To improve the performance of text candidate construction, Gomez et al. introduce grouping hypotheses based on different image features \cite{gomez2015object}. 

Compared to sliding window-based scene text detection, connected component analysis methods generate less number of extracted candidate components ensuring lower computational cost. However, each connected component analysis method considers hypotheses on the possible location of the text region candidate and these hypotheses fail to cover complex real-life scenarios.  

\subsubsection{Scene Text Detection Using Hybrid Approaches}
In the early deep learning era, statistical feature-based, machine learning-based, and deep learning-based techniques are applied in a mixed fashion to detect text in the natural scene. Convolutions neural network (CNN) is one of the most effective networks in deep learning while working with image data. Applying CNNs with the sliding window fashion, candidate text areas are identified and then the text location is predicted \cite{wang2012end}.  Jaderberg et al. acquire the text saliency map by utilizing CNNs with the sliding window and then apply the saliency map to forecast the boundary box of a word  \cite{jaderberg2014deep}. CNN is directly used to automatically extract features from the input image and the SVM classifier is applied to predict the text area \cite{kaur2022comprehensive}. Huang et al. apply CNN to extract robust and distinctive MSER features from the input image \cite{huang2014robust}.

In the hybrid approaches, the statistical features and machine learning algorithms are utilized in candidate text area generation or actual text location prediction. Therefore, the limitations of statistical feature-based or machine learning-based exist in hybrid approaches for scene text detection.

\subsubsection{Scene Text Detection Using Deep Learning-based Approaches}\label{subsection: detection with DL}
With the recent advent of deep learning-based techniques, a significant improvement has been made to solve computer vision tasks such as object detection and semantic segmentation. Inspired by deep learning-based techniques for general object detection and semantic segmentation, practitioners adopt related methods to solve the text detection from natural scene text. Deep learning-based techniques enable the automatic extraction of robust and highly distinctive features from the input image so that manual handcrafted feature engineering in machine learning-based approaches can be avoided. Moreover, automatically generated features are highly effective for real-life natural scenes imaginary with complex backgrounds. All the related works for scene text detection using deep learning-based approaches is divided into two different sub-categories:  (1) Scene text detection using segmentation techniques by considering a pixel-wise classification problem \cite{deng2018pixellink, long2015fully, yao2016scene, zhang2016multi, qin2017cascaded, long2018textsnake, wang2019shape, chen2019irregular, naosekpam2022utextnet, wang2019shape, yang2018inceptext, wang2019efficient, shao2021bdfpn, kobchaisawat2020scene}; and (2) Scene text detection using object detection techniques by considering a regression problem \cite{zhong2017deeptext, tian2016detecting, zhu2017deep, liao2017textboxes, liao2018textboxes++, huang2015densebox, gupta2016synthetic, naosekpam2021multi}. 

\subsubsection*{Scene Text Detection Using Segmentation Approaches}
Segmentation-based approaches address the scene text detection problem as a pixel-wise classification to identify the text and non-text segments in the given image. There are two different types of segmentation-based text detection approaches found in the literature: (1) Semantic segmentation-based \cite{deng2018pixellink, long2015fully, yao2016scene, zhang2016multi, qin2017cascaded, long2018textsnake, wang2019shape, chen2019irregular, naosekpam2022utextnet}; and (2) Instance segmentation-based \cite{wang2019shape, yang2018inceptext, wang2019efficient, shao2021bdfpn, kobchaisawat2020scene} techniques. 

Several scene text recognition techniques \cite{deng2018pixellink, long2015fully, yao2016scene, zhang2016multi, qin2017cascaded, long2018textsnake, wang2019shape, chen2019irregular, naosekpam2022utextnet} are introduced using semantic segmentation approaches. Using a Fully Convolutional Neural Network (FCNN) \cite{long2015fully}, the segmentation map is first generated from the input image and then the bounding boxes around the text segments are obtained with necessary post-processing. Yao et al. adapt the FCNN to produce three types of global score maps: one for character categories, another for text and non-text areas, and a third for linking orientation \cite{yao2016scene}. Moreover, a word partition method has been introduced to generate the bounding boxes around the text segments. More semantic segmentation approaches modifying FCNN has been found in \cite{ zhang2016multi, qin2017cascaded}. However, these methods show poor performance during bounding box prediction for closely adjacent words. To address the problem, PixelLinks \cite{deng2018pixellink} has been introduced to highlight the text margin. Moreover, TextSnake \cite{long2018textsnake} predicts text region and center lines to detect scene text. Another method called Progressive Scale Expansion Network (PSENet) \cite{wang2019shape} addresses the problem related to closely adjacent words by predicting different scale kernels. The attention-based mechanism has been utilized to detect the scene text using the semantic segmentation-based approach which helps to avoid the overlapping closely adjacent text \cite{chen2019irregular}. Using an encoder-decoder neural network, UtextNet \cite{naosekpam2022utextnet} introduces a scene text recognition model utilizing the UNet-ResNet50 and a post-processing technique to predict the bounding boxes.

In the literature, we have found several techniques \cite{wang2019shape, yang2018inceptext, wang2019efficient, shao2021bdfpn, kobchaisawat2020scene} that consider scene text detection as a similar problem comparing instance segmentation. SPCNET \cite{wang2019shape} utilizes the Mask RCNN \cite{he2017mask} model architecture to design the text context module and a re-score mechanism to improve the scene text detection. Inceptext \cite{yang2018inceptext} introduces an instance-aware segmentation mechanism to enhance the text detection performance for large-scale text. By using a feature pyramid enhancement module and feature fusion module, an improved version of the arbitrary text detection model has been proposed in \cite{wang2019efficient}. Shao et al. propose a bi-directional feature pyramid network to improve text detection performance in blurry and low-contrast images \cite{shao2021bdfpn}. Border augmentation using a combination of polygon offsetting has been utilized in \cite{kobchaisawat2020scene} to detect scene text. However, segmentation-based text detection models require high inference time due to large model sizes and are less suitable for real-time text detection.

\subsubsection*{Scene Text Detection Using Object Detection Techniques}
By assuming the text as an object, researchers acknowledge that scene text detection is a sub-problem of the general object detection problem. Therefore, scene text detection is considered a regression problem to predict bounding boxes around the text region. We categorize text detection into 2 different groups: (1) Two-stage detection approaches \cite{zhong2017deeptext, tian2016detecting, zhu2017deep}; and (2) One-stage detection approaches \cite{liao2017textboxes, liao2018textboxes++, huang2015densebox, gupta2016synthetic, naosekpam2021multi}.

In the two-stage detection approach, the text region proposals are first generated by selective search and then applied classification to each text region proposal. R-CNN \cite{girshick2014rich} and its variants such as Fast R-CNN \cite{girshick2015fast}, Faster R-CNN \cite{ren2015faster}, and Mask R-CNN \cite{he2017mask} are the popular two-stage object detection approaches among researchers. Zhong et al. propose a text detection model named DeepText by modifying R-CNN \cite{zhong2017deeptext}. Tian et al. utilize the Fast R-CNN, an improvement of R-CNN, to design the Connectionist Text Proposal Network (CTPN) for text detection \cite{tian2016detecting}. Zhu et al. propose an improvement of CTPN by incorporating a vertical proposal mechanism \cite{zhu2017deep}. However, these two-stage detection approaches are comparatively slow and are not applicable to real-time text detection scenarios.

One-stage detection approaches incorporate a single network for both region proposal generation and region proposal classification. SSD \cite{liu2016ssd}, EAST \cite{zhou2017east}, Yolo \cite{redmon2016you} and its improved variations such as  YoloV2 \cite{redmon2017yolo9000}, YoloV3 \cite{redmon2018yolov3}, YoloV4 \cite{bochkovskiy2020yolov4}, and YoloV5\footnote{https://github.com/ultralytics/yolov5} are the popular one-stage object detection approaches found in the literature. Liao et al. propose a text detection model named Textboxes by modifying the Single-Shot Detection (SSD) kernel \cite{liao2017textboxes}. Another improved version of Textboxess++ \cite{liao2018textboxes++} considers the predicted text region as quadrilateral in place of rectangular bounding boxes. Densebox \cite{huang2015densebox} is an improved version of the previously proposed EAST \cite{zhou2017east} where a fully connected network has been employed to detect text scores and geometry at the pixel level. Based on the Yolo-based object detection model, a fully convolutional regression network has been proposed to detect text in an image \cite{gupta2016synthetic}. Naosekpam et al. introduce Yolov3 and Yolov4-based shallow networks to detect multi-lingual scene text \cite{naosekpam2021multi}. Yolo-based models for scene text detection outperform all the previously proposed models.

\subsection{Scene Text Recognition}
Text recognition from images is a fundamental research problem in the field of computer vision. A significant amount of effort has been devoted to recognizing text from scanned document images. Researchers have achieved incredible accuracy to recognize the text from scanned document images for resource-enriched languages like English and a significant number of optical character-recognizing applications have been emerging for practical usage. However, recognizing text from natural scenes is more challenging than recognizing text from scanned documents using Optical Character Recognizer (OCR). Natural scene images have a complex background with inconsistent and irregular font style, font size, and multi-color text. Moreover, noise, blur, distortion, and skewness are the common characteristics of natural scene imagery. Therefore, existing OCRs fail to recognize text from natural scenes.  In the literature, we have found several previous research works which address natural scene text recognition. The related previous approaches for scene text recognition can be divided into two major categories: (1) Scene text recognition using classical machine learning-based approaches \cite{chen2004automatic, de2009character, wang2010word, sheshadri2012exemplar, ali2016character}; and (2) Scene text recognition using deep learning-based approaches \cite{isthiaq2020ocr, bissacco2013photoocr, jaderberg2016reading, graves2007unconstrained, he2016reading, bai2018edit, liu2018squeezedtext}.

\subsubsection{Scene Text Recognition Using Machine Learning-based Approaches}
In classical machine learning approaches, manually handcrafted features are extracted from the input image and then scene texts are recognized using these extracted features. The text recognition problem can be decomposed into a series of different sub-problems such as text binarization, segmentation of individual lines of text, segmentation of individual characters, recognition of individual characters, and finally merging them into full text.  Several different methods have been found in the literature for the each of sub-problems such as text binarization \cite{zhou2010edge, mishra2011mrf, lee2013integrating},  segmentation of individual line of text \cite{ye2003robust}, segmentation of individual characters \cite{nomura2005novel, shivakumara2011new, roy2009multi}, recognition of individual characters \cite{chen2004automatic, sheshadri2012exemplar} and finally merging them into full text \cite{weinman2007fast, mishra2012scene}. 

Statistical features such as SIFT \cite{lowe2004sift} and HOG \cite{dalal2005histograms} are extracted, SVM classifies each character and is followed by post-processing steps to generate the final text by merging individual characters. Wang and Belongie adapt the HOG features and deploy Nearest Neighbor (NN) for character classification \cite{wang2010word}. Campos et al. conduct object categorization using a bag-of-visual-words (BoW) representation for text recognition \cite{de2009character}.  More feature extraction techniques such as tensor decomposition \cite{ali2016character}, shape context \cite{belongie2002shape}, or patch descriptors \cite{varma2002classifying} have been utilized for character classification.

Machine learning approaches require an intensive and complex pipeline of pre-processing steps to extract manually handcrafted low-level or mid-level potential features for character classification and post-processing steps to generate full text from the series of recognized characters. However, due to the limited representation constraints of traditional machine learning methods and the high complexity of pre-processing and post-processing pipelines, the machine learning-based text recognition methods hardly deal with the challenging characteristics found in natural scene imagery and show poor performance in scene text recognition.
\subsubsection{Scene Text Recognition Using Deep Learning-based Approaches} \label{subsection: recognition with DL}
Deep learning-based approaches facilitate automatic feature extraction from the input image. Convolutional Neural Network (CNN) based deep learning framework extracts high-level features automatically from natural scene images with complex characteristics. We can divide deep learning-based approaches into two main categories: (1) Segmentation-based scene text recognition \cite{wang2012end, sen2022end, ahmed2019complete, isthiaq2020ocr, bissacco2013photoocr, jaderberg2016reading}; and (2) Segmentation-free scene text recognition \cite{graves2007unconstrained, he2016reading, shi2016end, gao2017reading, lee2016recursive, cheng2017focusing, bai2018edit, liu2018squeezedtext}. Moreover, segmentation-free techniques utilize two different frameworks for scene text recognition: (1) CTC-based framework \cite{graves2007unconstrained, he2016reading, shi2016end, gao2017reading}; and (2) Encoder-Decoder framework \cite{lee2016recursive, cheng2017focusing, bai2018edit, liu2018squeezedtext}. 
\subsubsection*{Segmentation-based Scene Text Recognition}
Segmentation-based scene text recognition requires a preprocessing pipeline to segment the text line, segment individual characters, then recognize each character with a character classifier, and finally pro-processing step to generate the final text.

Wang et al. introduce a CNN-based model architecture to classify each character and provide Non-maximal Suppression (NMS) algorithm to generate the final text \cite{wang2012end}. Sen et al. propose a U-net-based model architecture with a matra-removal strategy to recognize Bangla and Devanagari characters \cite{sen2022end}. More character recognition models are found in \cite{ahmed2019complete, isthiaq2020ocr, bissacco2013photoocr, jaderberg2016reading}. However, each of the segmentation-based scene text recognition techniques requires localizing the individual character which is a challenging problem due to the complex characteristics of the natural scene image. Therefore, segmentation-based scene text recognition techniques provide poor performance for scene text recognition. 

\subsubsection*{CTC-based Framework for Segmentation-free Scene Text Recognition}
The Connectionist Temporal Classification (CTC) decoding module originates from a speech recognition system. The CTC-based framework deals with sequential data. While applying a CTC-based framework for image data, we have to consider the input image as a sequence of frames of vertical pixels. Then, the CTC rules are applied to generate the target sequence from the per-frame prediction. Therefore, the CTC-based framework provides an end-to-end trainable network for text recognition. 

The first attempt to apply the CTC-based framework for the OCR system is found in \cite{graves2007unconstrained}. The CTC-based framework is widely incorporated to solve the scene text recognition problem \cite{he2016reading, shi2016end, gao2017reading}. The Convolutional Recurrent Neural Network (CRNN) is designed for a sequence of feature slice generation by including RNN layers after the stack of CNN layers and finally using CTC rules to generate the target sequence. He et al. introduce DTRN model using CRNN and CTC loss \cite{he2016reading}. Shi et al. modify the DTRN model by introducing a fully convolutional approach to generate the sequence of feature slices \cite{shi2016end}. Gao et al. replace  RNN layers by adapting the stacked CNN layers to generate the feature slices \cite{gao2017reading}. 

\subsubsection*{Encoder-Decoder Framework for Segmentation-free Scene Text Recognition}
An encoder-decoder framework is a popular approach for sequence-to-sequence learning problems \cite{sutskever2014sequence}. The encoder network takes the sequence data as input and creates the final latent state. Using the latent state as input the decoder network generates the output sequence in an auto-regressive manner. The encoder-decoder framework is very effective when the length of the output is variable, which is the requirement for the scene text recognition problem. Moreover, the performance of the encoder-decoder framework is improved by adapting the attention mechanism to jointly learn to align the input and output sequence properly \cite{bahdanau2014neural}. 

A recursive recurrent neural network with an attention mechanism has been introduced in \cite{lee2016recursive} where the encoder network with a recursive convolutional layer learns the feature vector from the input image, then the attention layer ensures the best feature selection, and finally, the decoder network with recurrent neural network generates the character sequence.  Cheng et al. propose an improved version of the attention mechanism by imposing localization supervision while calculating the attention score \cite{cheng2017focusing}. Moreover, an improved attention mechanism has been proposed to minimize misalignment problems in output sequence \cite{bai2018edit} and reduce the computational cost \cite{liu2018squeezedtext}.

Segmentation-free scene text recognition approaches with encoder-decoder and CTC-based framework simplify the recognition pipeline by drastically eliminating the complex preprocessing and post-processing steps and enable training the model without character level annotated dataset. The encoder-decoder and CTC-based frameworks require a larger dataset to train the model architecture, which is now possible by creating a large synthetic dataset \footnote{https://github.com/Belval/TextRecognitionDataGenerator}.

\subsection{Address Parsing}
Automatic address parsing is an active research area with real-life practical applications such as efficient searching on the mapping platform, or automatic address insertion on a relational dataset. It is a challenging problem due to the large variety of user address input even in the same language. There are two different approaches for automatic address parsing: (1) Address parsing using machine learning-based approaches \cite{borkar2001automatic, churches2002preparation, li2014hmm, wang2016probabilistic}; and (2) Address parsing using deep learning-based approaches \cite{mokhtari2019tagging, craig2019scaling, yassine2021leveraging}.

\subsubsection{Address Parsing Using Machine Learning-based Approaches}
Traditional rule-based machine-learning methods have been proposed for automatic address parsing in \cite{borkar2001automatic, churches2002preparation}. However, it has been found that rule-based methods require prior domain knowledge which is not available due to the complex domain of possible address. To improve the performance of rule-based methods, probabilistic methods based on the Hidden Markov Model (HMM) and Conditional Random Field (CRF) have been introduced in \cite{li2014hmm, wang2016probabilistic}. Large-scale HMM-based parsing techniques are capable of working with large variations than rule-based methods \cite{li2014hmm}. A linear-chain CRF combined with a Stochastic Regular Grammar (SRG) is utilized to create a discriminative model to deal with the complex domain of possible user address input \cite{wang2016probabilistic}.  However, probabilistic methods based on HMM and CRF heavily rely on structured data which is not available due to the variety of user input.

\subsubsection{Address Parsing Using Deep Learning-based Approaches}
In recent years, the deep learning-based neural network has been utilized to solve the automatic address parsing problem. Sharma et al. propose a multi-layer feed-forward neural network that provides better results than rule-based or probabilistic models \cite{sharma2018automated}. RNN-based models utilize the sequential properties of the address text and provide state-of-the-art results for automatic address parsing. Mokhtari et al. propose several sequence-to-sequence models with RNN, Bi-RNN, LSTM, or Bi-LSTM units and conduct a comparative study among them for automatic address parsing \cite{mokhtari2019tagging}. Several sequence-to-sequence models with RNN and its variations have been found in the literature \cite{craig2019scaling, yassine2021leveraging}. 
 
\begin{figure*}[h!]
  \centering
  \includegraphics[width=\textwidth]{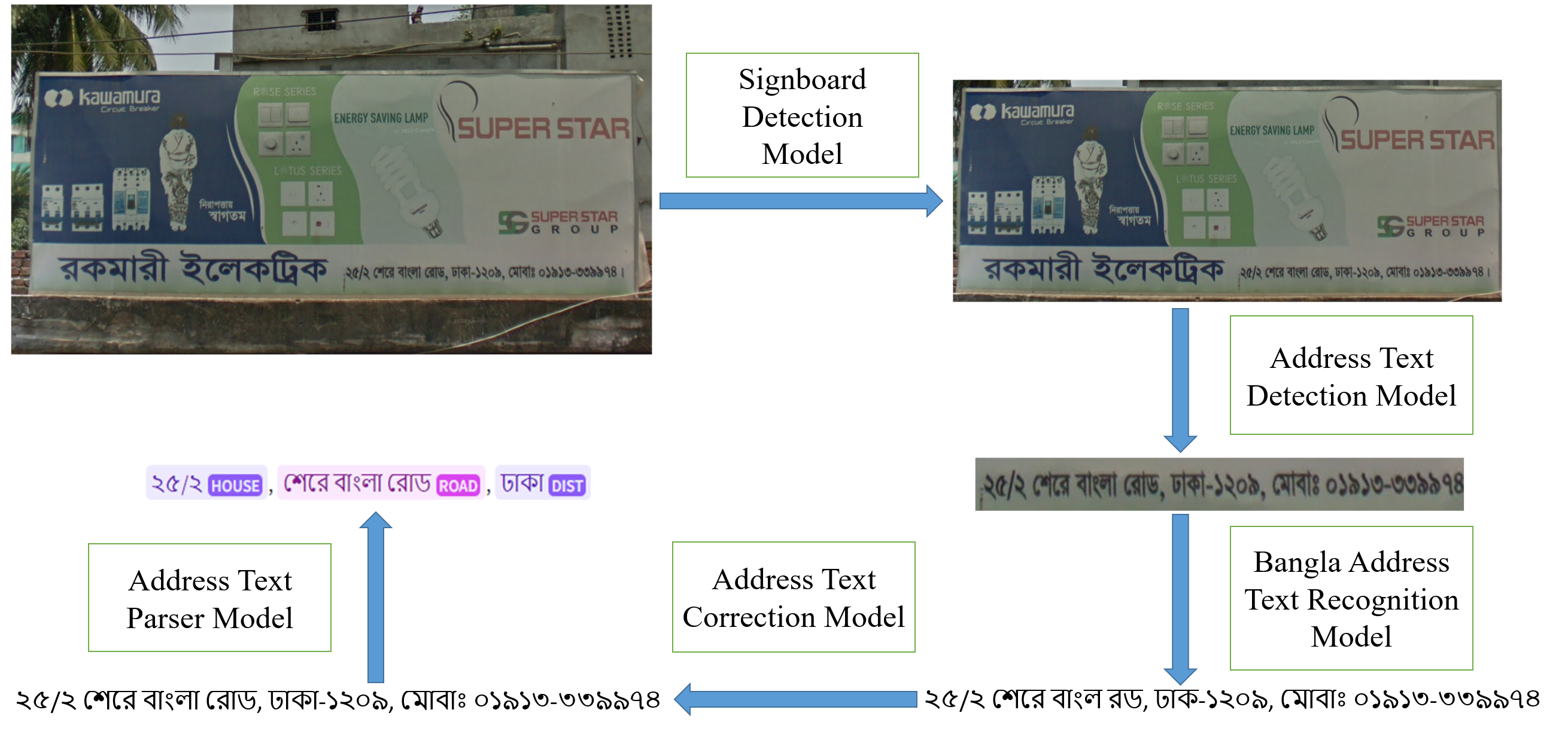}
  \caption{An overview of the end-to-end system}
  \label{fig:soluton overview 1}
\end{figure*}
\section{Methodology}\label{methodology}

In this section, we discuss the methodology of our research work. To develop an end-to-end system for extracting and parsing address information from the signboard, we have divided the whole system into different sub-problems: (1) a signboard detection model for detecting signboard from the raw image, (2) an address text detection model for detecting address portion from a signboard, (3) a Bangla address text recognition model to extract address text from the cropped address image,  (4) an address text correction model to improve the output of the Bnagla address text recognition model, and finally, (5) an address text parser model to classify each field of an address. Figure \ref{fig:soluton overview 1} shows an overview of the end-to-end system. We discuss the possible model architectures used to train and evaluate the model for each sub-problem.

\subsection{Detection Models}
In this research, we have developed two detection models - the signboard detection model and the address text detection model. The signboard detection model detects the signboard portion from the natural scene image and the address text detection model then detects the address text portion from the cropped signboard image. 

The deep learning-based approaches are the most effective technique for detection. There are two different deep learning-based approaches for detection such as segmentation-based and object detection based where object detection-based techniques are the most effective one for real-time detection. There are two different types of object detection approaches such as two-stage detection and one-stage detection where one stage detection approach is faster and more accurate than two-stage detection.

\begin{figure*}[h!]
  \centering
  \includegraphics[width=\textwidth]{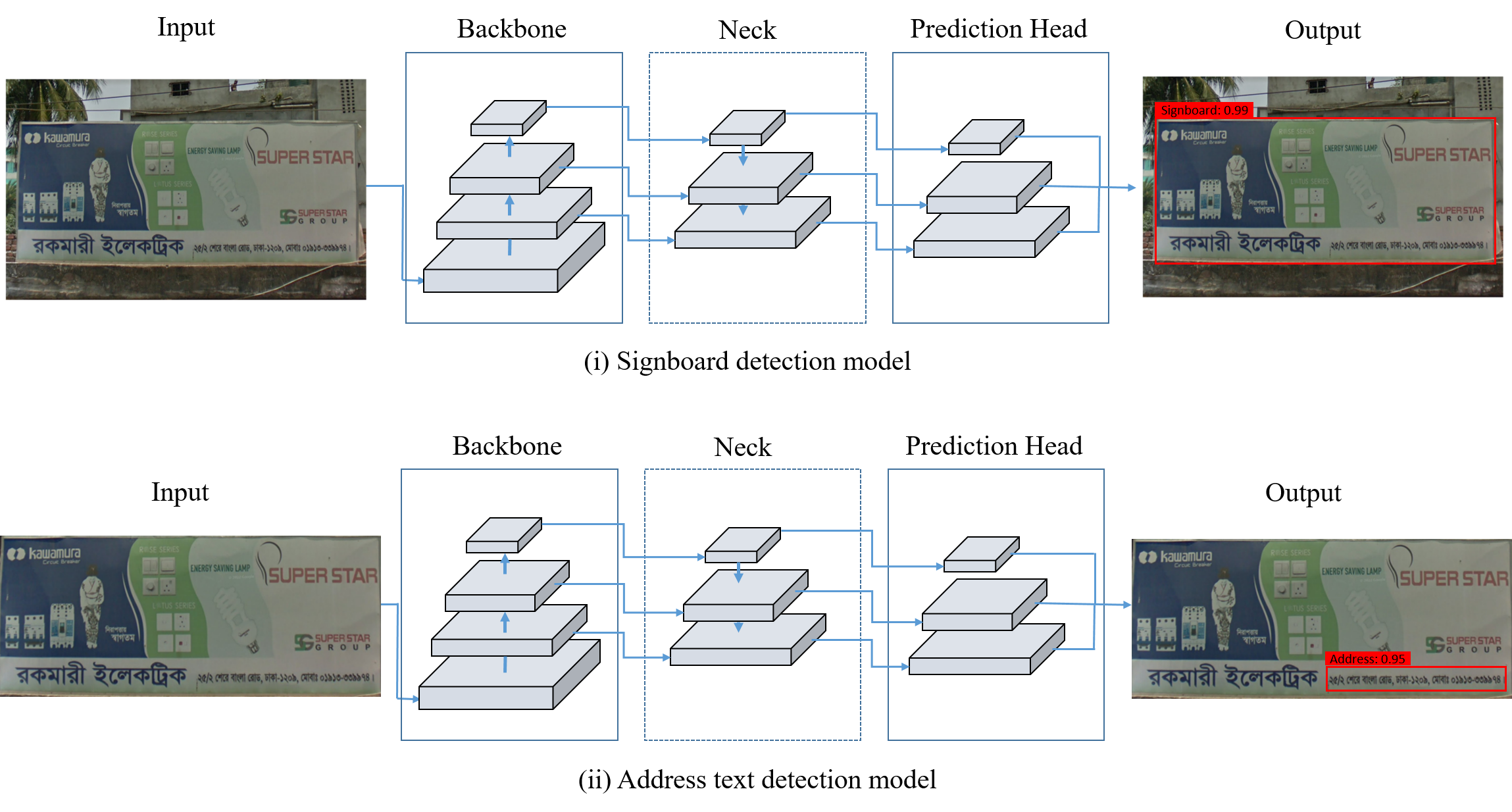}
  \caption{A high-level overview of the Yolo-based model architecture for both the signboard detection model and the address text detection}
  \label{fig: detection model}
\end{figure*}

The Yolo-based object detection model and its improved versions are examples of one-stage detection approaches. In this research work, we train and evaluate different recent Yolo-based object detection models such as YOLOv3, YOLOv4, and YOLOv5 for both the signboard detection model and the address text detection model. Figure \ref{fig: detection model} shows a high-level overview of the Yolo-based model architecture for both the signboard detection model and the address text detection model.

The Yolo-based model is a single unified network that considers the object detection problem as a regression problem to simultaneously predict the rectangular bounding boxes and the class-wise probabilities for each object. The Yolo-based model enables to design and train end-to-end models for real-time object detection. There are three main components of the Yolo-based model: backbone, neck, and prediction head. 

\subsubsection*{Backbone}
The initial component of the Yolo-based model is referred to as the backbone, and its primary role is to extract features from the input image. This backbone is constructed using deep-stacked convolutional neural networks. In deep-stacked convolutional neural networks, the first layers perform the task of extracting low-level features from the entire input image. As for the subsequent layers, they focus on extracting high-level features by utilizing the low-level features obtained from the preceding convolutional layer.

\subsubsection*{Neck}
The neck, as the second element of the YOLO-based model, plays a crucial role in extracting features necessary for detecting objects of various sizes. The neck consists of pooling layers and each pooling layer has a different kernel size. By taking output features from multiple layers of the backbone, the neck generates high-level feature maps. This feature map enables the YOLO-based model to effectively handle variations in scale among objects.

\subsubsection*{Prediction Head}
The final part of the Yolo-based model is the prediction head. It consists of different detection layers. Each detection layer is responsible for predicting the rectangular bounding boxes and the class-wise probabilities at different scales and aspect ratios. The head of the Yolo-based model consists of convolution layers and fully connected layers.

The YOLOv3 model uses the Darknet-53  architecture as the backbone model for feature extraction from the input image. The feature fusion layers are utilized to design the neck of the YOLOv3 model.  The feature fusion layers apply upsampling and concatenation operations to combine the features from different layers of Darnet-53 architecture so that the YOLOv3 model can handle the scale variation of the signboard. The head of YOLOv3 consists of different detection layers to predict the rectangular bounding boxes and the class-wise probabilities.

The YOLOv4 model utilizes a modified version of the CSPDarknet53 architecture as the backbone, which extracts high-level features from the input image through stacks of convolutional layers. The neck component of YOLOv4 employs feature fusion layers and feature pyramid modules to combine features from different scales produced by different layers of the backbone. This fusion allows the model to effectively handle variations in scale. Additionally, the neck incorporates additional convolutional layers to refine and integrate the features further. The head of the YOLOv4 model contains multiple detection layers responsible for predicting rectangular bounding boxes and class probabilities.

The YOLOv5 model also uses the CSPDarknet53 architecture as the backbone. The YOLOv5 model utilizes the spatial pyramid pooling module with Path Aggregation Network(PANet) to design the neck. Finally, the head of YOLOv5 consists of different detection layers to predict the rectangular bounding boxes and the class-wise probabilities. The implementation of the YOLOv5 model is on a popular Python library called Pytorch so that researchers can develop an object detection model easily compared to the previous version of the Yolo-based model. The performance of the YOLOv5 model is almost similar to the YOLOv4 model as both models use the same backbone of CSPDarknet53.

In this research work, we train and evaluate different recent Yolo-based object detection models such as YOLOv3, YOLOv4, and YOLOv5 for both the signboard detection model and the address text detection model using the Bangla Signboard Detection Dataset and Bangla Address Detection Dataset respectively. 

\begin{figure*}[h!]
  \centering
  \includegraphics[width=\textwidth]{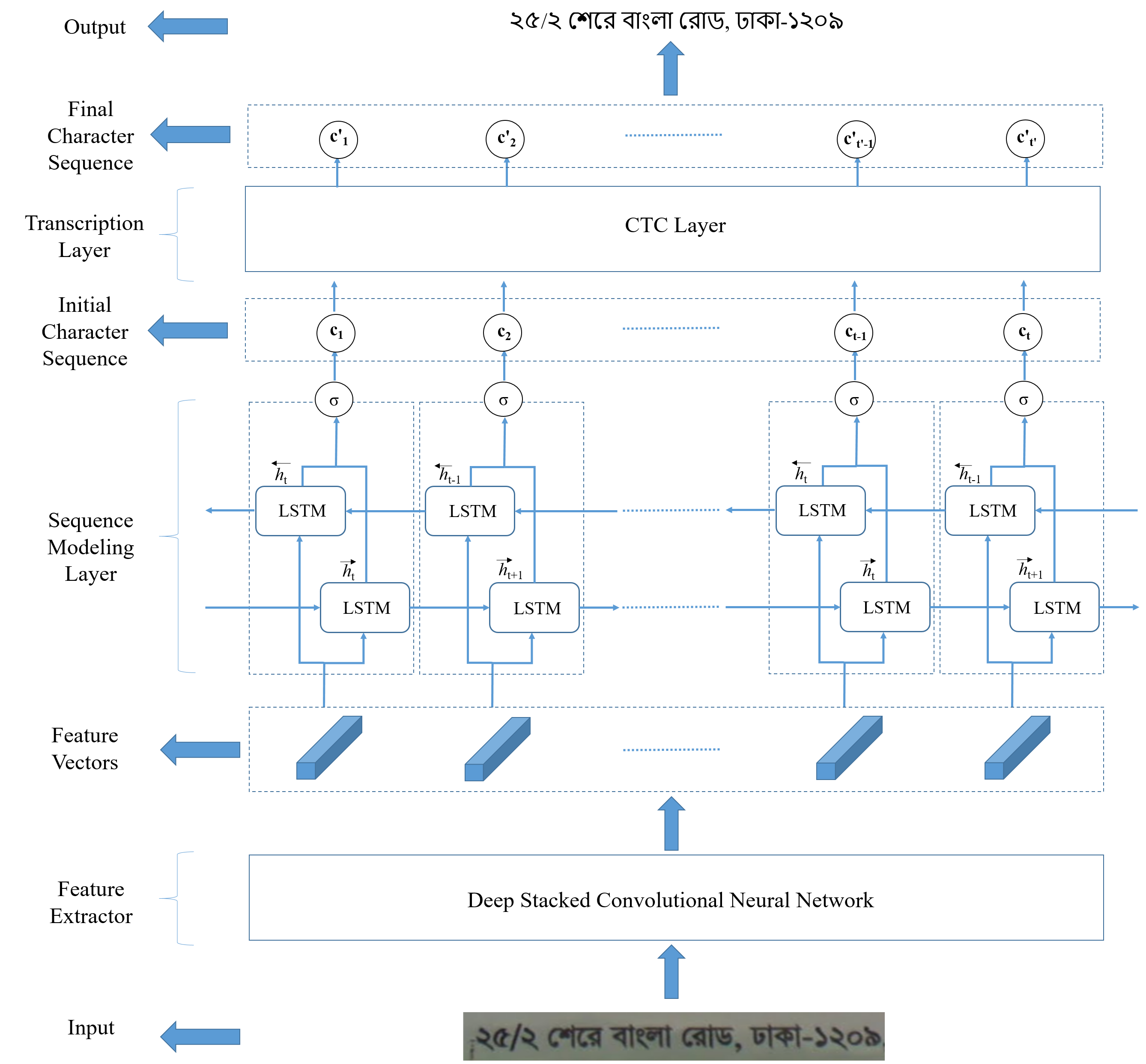}
  \caption{The CTC-based Model Architecture for Bangla Address Text Recognition}
  \label{fig: Bangla Text Recognition Model Architecture}
\end{figure*}

\subsection{Bangla Address Text Recognition Model}
After detecting the address text portion from the signboard, the next step is to recognize the address text from the cropped address text portion image. From Subsection \ref{subsection: recognition with DL}, we have come to know that there are two different deep learning-based approaches for scene text detection such as segmentation-based and segmentation-free techniques where the segmentation-free technique is more effective to train an end-to-end text recognition model. For scene text recognition for segmentation-free deep learning-based approaches, there are two different frameworks such as the CTC-based framework and the Encoder-Decoder framework. In this research work, we train and evaluate different recent CTC-based and Encoder-Decoder models for Bangla address text recognition.

\subsubsection{CTC-based Model Architecture for Bangla Address Text Recognition}

For Bangla address text recognition, we have designed the CTC-based model which consists of three major components: (1) Convolutional layers for feature extraction; (2) Recurrent layer for sequence modeling; (3) Transcription layer with CTC for sequence prediction. Figure \ref{fig: Bangla Text Recognition Model Architecture} shows an overview of the CTC-based model architecture for Bangla address text recognition.

\subsubsection*{Convolutional Layers for Feature Extraction}
Convolutional layers are the popular and most effective automatic feature extractor for visual recognition tasks. Before feeding the input image into convolutional layers, we need to rescale the image to the same height. The convolutional layers generate feature vectors from the input image. Each feature vector represents feature maps generated from each receptive field which is a rectangle region on the given image. In this research work, we have utilized different deep-stacked CNN architectures such as VGG, RCNN, GRCL, and ResNet for feature extraction.

\subsubsection*{Recurrent Layer for Sequence Modeling}
We use a bidirectional recurrent layer on top of the previous convolutional layers for sequence modeling. The recurrent layer predicts the sequence of labels from feature vectors generated by the convolutional layers for rectangle regions. There are many advantages of using a recurrent layer for sequence modeling. We are generating a sequence of characters from the input image and the recurrent layer has a strong capability to address the contextual information of the sequential data. Using contextual information, the recurrent layer helps to generate the next character based on both the current feature vector and the previously predicted character sequence. The recurrent layer can deal with sequences of arbitrary length. Moreover, the recurrent layer can be trained with previous convolutional layers as a single neural network. To design the recurrent layer, there are different types of recurrent units such as RNN, Bi-RNN, LSTM, or Bi-LSTM. RNN fails to address the contextual information of the long sequence while LSTM has the capability to capture the contextual information on the long sequence. Moreover, the bidirectional recurrent layer is more effective than the unidirectional recurrent layer as the bidirectional recurrent layer can address the contextual information from both directions on the sequence. Therefore, in this research work, we design the recurrent layer using the Bi-LSTM units for sequence modeling. The recurrent layer generates a sequence of initial characters from the feature vectors where each character can be mapped with a feature vector and each feature vector represents a receptive field in the input image. 

\subsubsection*{Transcription Layer with CTC for Sequence Prediction}
The return sequence of the recurrent layer contains a significant number of repeated characters as multiple receptive fields can be possible on the width of a single character in the input image. For example, ``--hh-eee-ll--lll--oo-'' is generated by the recurrent layers for ``hello''. We utilize the conditional probability defined by the CTC layers to remove the repetitive characters and generate the target sequence. Suppose, the predicted sequence of the recurrent layers is $y = y_1, . . . , y_T$ where T is the length of the sequence and $l$ is the target sequence. $\beta$ is a sequence-to-sequence mapping function from the output sequence of the recurrent layers to the target sequence. We calculate the conditional probability $l$ given $y$ using the following Equation \ref{equation: CTC}.

\begin{equation}
    p(l|y) = \sum_{\pi:\beta(\pi)=l} p(\pi|y)
    \label{equation: CTC}
\end{equation}
Where $\pi \in L^T$ and $L^T$ is the set of all possible target sequences. 

The CTC layer learns the CTC rules to remove the repeated characters from the output of the recurrent layer to generate the final character sequence.


In this research work, we design different Bangla address text recognition models using the CTC-based approach: VGG+Bi-LSTM+CTC, RCNN+Bi-LSTM+CTC, ResNet+Bi-LSTM+CTC, and GRCL+Bi-LSTM+CTC. We train and evaluate each model using the Syn-Bn-OCR Dataset and find out the best-performing model.

\begin{figure*}[!h]
  \centering
  \includegraphics[width=\textwidth]{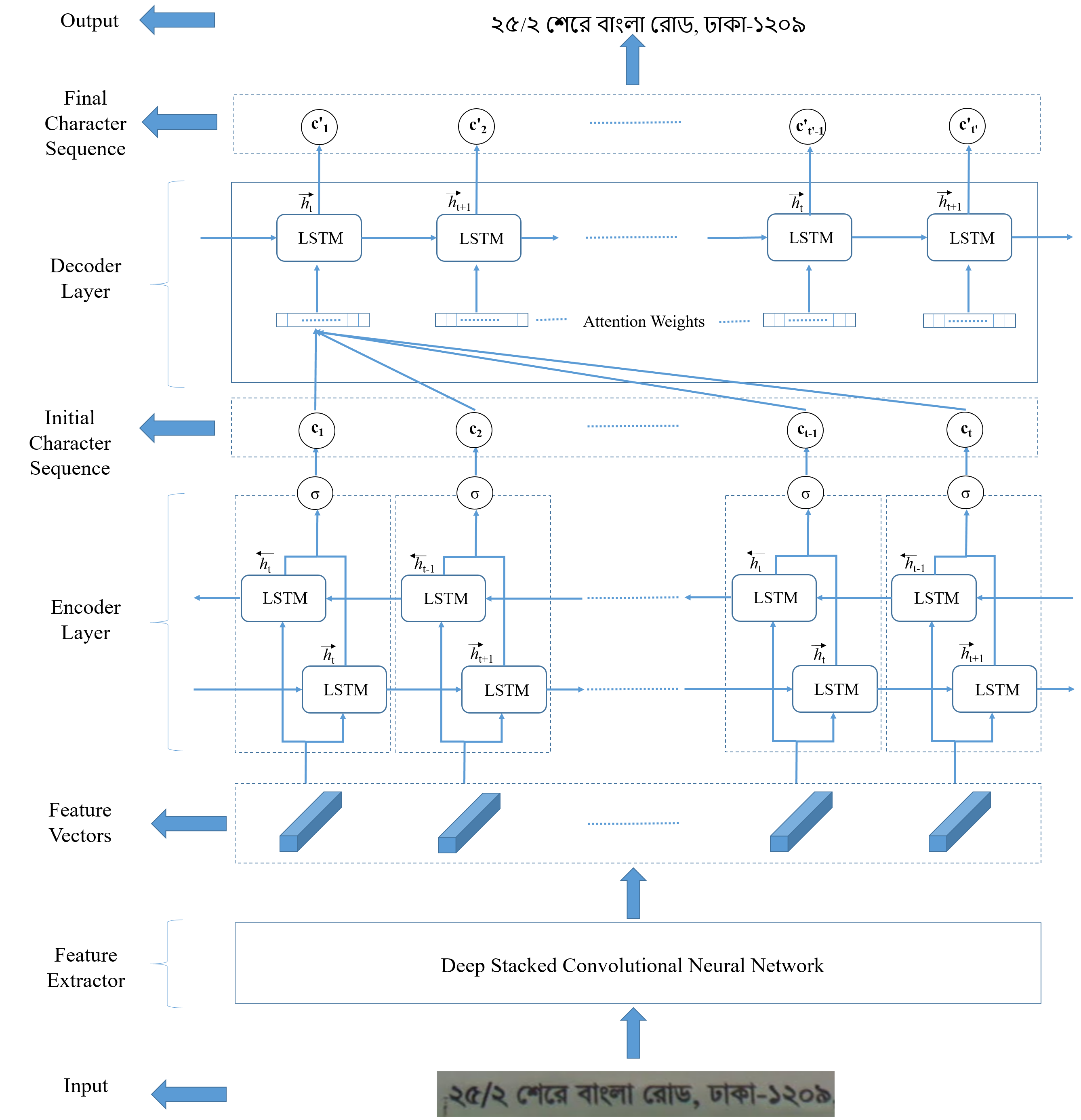}
  \caption{The Encoder-Decoder Model Architecture for Bangla Address Text Recognition}
  \label{fig: encoder-decoder model}
\end{figure*}

\subsubsection{Encoder-Decoder Model Architecture for Bangla Address Text Recognition}
 We have designed an Encoder-Decoder model architecture for the Bangla address text recognition model. Figure \ref{fig: encoder-decoder model} shows the proposed Encoder-Decoder model for Bangla address  text recognition. The Encoder-Decoder model architecture consists of three key components: (1) A deep-stacked convolution neural network for feature extraction; An encoder network with a recurrent layer for sequence modeling; and A decoder network with an attention mechanism for transcription. 

\subsubsection*{Convolution Layers}
Similar to the CTC-based model architecture, we use deep-stacked convolutional layers for the feature extraction from the input image. The convolutional layer generates feature maps for each receptive field on the input image. The feature map is the input to the next encoder network for sequence modeling. We have utilized different deep-stacked CNN architectures such as VGG, RCNN, GRCL, and ResNet.
\subsubsection*{Encoder Network}
The encoder network consists of recurrent layers which are responsible for sequence modeling. From different recurrent layers, we use the Bi-directional LSTM (Bi-LSTM) which is more effective to capture the contextual information of the input from both directions. Moreover, the Bi-directional recurrent layer can handle long sequences effectively compared to the unidirectional recurrent layer. The output of the encoder network is an intermediate representation which will be the input to the decoder network. 
\subsubsection*{Decoder Network}
The decoder network consists of an attention layer followed by an LSTM layer. In the attention layer, the input vector is multiplied with an attention weight vector before providing it as input to each LSTM cell. The attention weight vector is learned during the training phase. The output of the decoder layer is the final sequence of characters.

In this research work, we design different Bangla address text recognition models using the Encoder-Decoder approach with attention mechanism: VGG+Bi-LSTM+Attention, RCNN+Bi-LSTM+Attention, ResNet+Bi-LSTM+Attention, and GRCL+Bi-LSTM+Attention. We train and evaluate each model using the Syn-Bn-OCR Dataset and find out the best-performing model.

\begin{figure*}[!h]
  \centering
  \includegraphics[width=\textwidth]{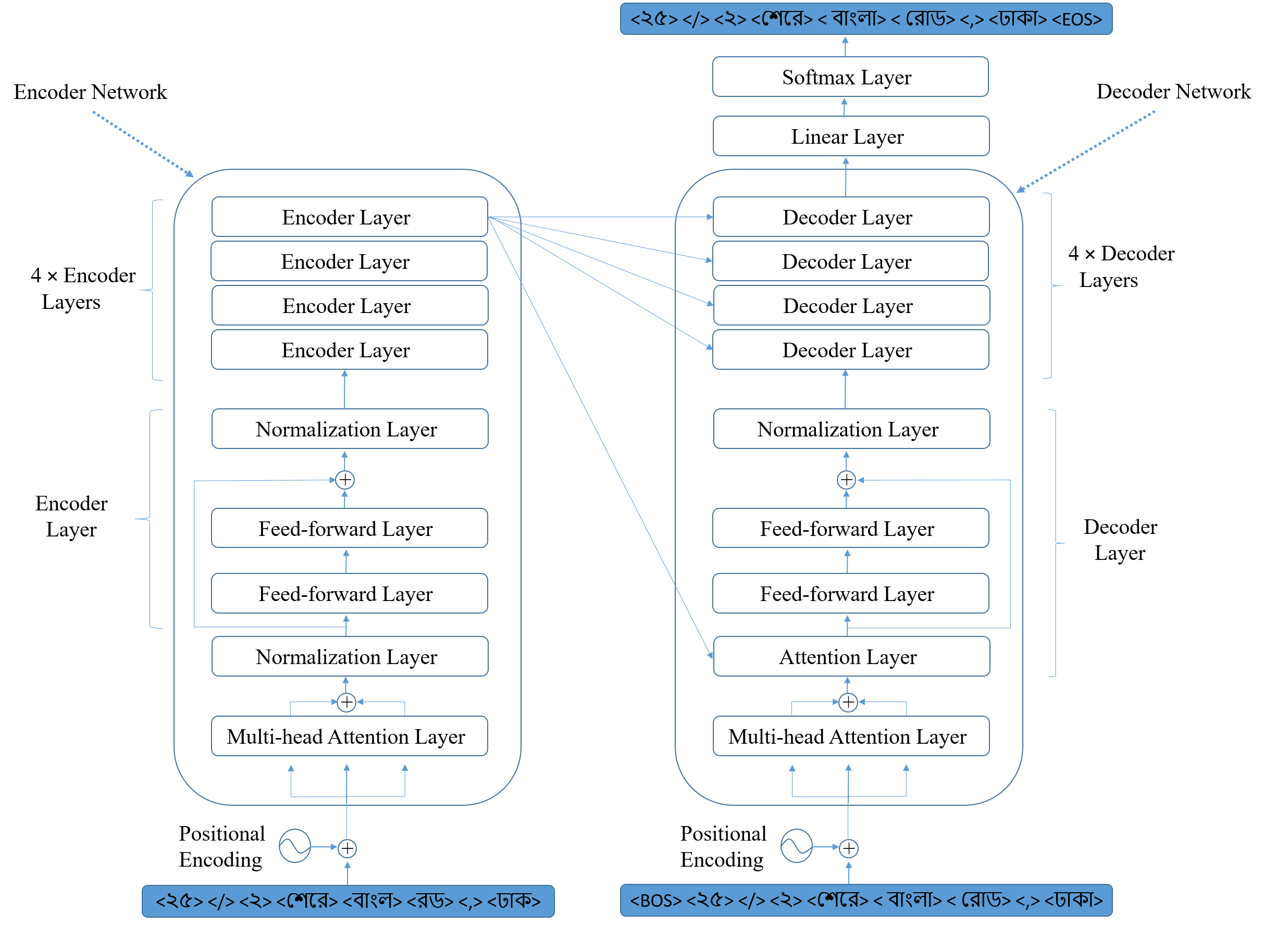}
  \caption{The transformer-based encoder-decoder model architecture for Bangla address text correction}
  \label{fig: Bangla Address Correction Model}
\end{figure*}

\subsection{Address Text Correction Model}
To improve the performance of the Bangla address text recognition by post-correction, we propose an address text correction model to automatically correct the output address text sequence of the Bangla address text recognition model. Due to the complexity of the scene text recognition, the output character sequence of the Bangla address text recognition model sometimes contains wrong characters which can be corrected by the contextual information of the character sequence. The sequence-to-sequence Encoder-Decoder framework is appropriate for the address text correction model.

Address text correction is a sequence-to-sequence modeling problem. The encoder-Decoder framework is the most effective model architecture for sequence-to-sequence modeling problems. For Bangla address text correction, we have designed a transformer-based encoder-decoder model inspired by the Bangla to English translation model \footnote{https://huggingface.co/Helsinki-NLP/opus-mt-bn-en} found in \cite{tiedemann2020opus}. Figure \ref{fig: Bangla Address Correction Model} shows an overview of the transformer-based encoder-decoder model. 

The transformer model comprises two essential components: the encoder network and the decoder network. These networks consist of various crucial elements, including positional encoding, multi-head attention layer, feed-forward layer, as well as residual connections and layer normalization. 
\subsubsection*{Positional Encoding}
The transformer model does not employ convolutional or sequential operations on the input sequence, instead relying on position encoding to account for the sequential order of the input data. By adding positional encodings to the embedded input tokens, the transformer model is able to determine the relative position of each token within the embedding.

\subsubsection*{Multi-head Attention Layer}
The multi-head attention layer consists of multiple self-attention layers that work simultaneously. Self-attention, also called Scaled Dot-Product Attention, is a mechanism intended to capture relationships within a sequence. Self-attention enables the model to effectively incorporate long-term dependencies. The multi-head attention layers employ multiple attention heads in parallel to acquire complex pattern within the input sequence easily. 

\subsubsection*{feed-forward layer}
The feed-forward layer utilizes both a linear transformation and a non-linear activation function. By incorporating the feed-forward layer, the transformer model introduces non-linearity, enabling it to learn complex patterns and complex relationships within the sequence.

\subsubsection*{Residual Connections and Layer Normalization}
To overcome the vanishing gradient problem and enhance information flow, the network includes residual connections. These connections are followed by layer normalization, which stabilizes the distribution of hidden states, leading to accelerated training of the transformer model. Layer normalization helps mitigate the negative effects of input variations and internal covariate shift, reducing their influence on the model.

\subsubsection*{Encoder Network}
The encoder network of the transformer layer consists of a multi-head attention layer followed by encoder layers. The input embedding is added with the positional encoding before providing it as the input to the multi-head attention layer. Each encoder layer consists of two feed-forward layers followed by a normalization layer. Moreover, there is a residual connection from the output of the previous encoder layer to the input of the normalization layer. The encoder network generates the encoded context from the incorrect address text.
\subsubsection*{Decoder Network}
The decoder network consists of a multi-head attention layer followed by decoder layers. Each decoder layer consists of an encoder-decoder attention layer, two feed-forward layers, and a normalization layer. Each decoder layer takes both the output of the previous layer and the context vector of the encoder layer as input. The output of the decoder network is passed as the input to a linear layer. Using the encoded context, the decoder network generates the corrected address text. 

In this research work, we train and evaluate the transformer-based encoder-decoder model using the synthetically generated Bangla address correction (Syn-Bn-AC) dataset.

\begin{figure*}[!h]
  \centering
  \includegraphics[width=0.8\textwidth]{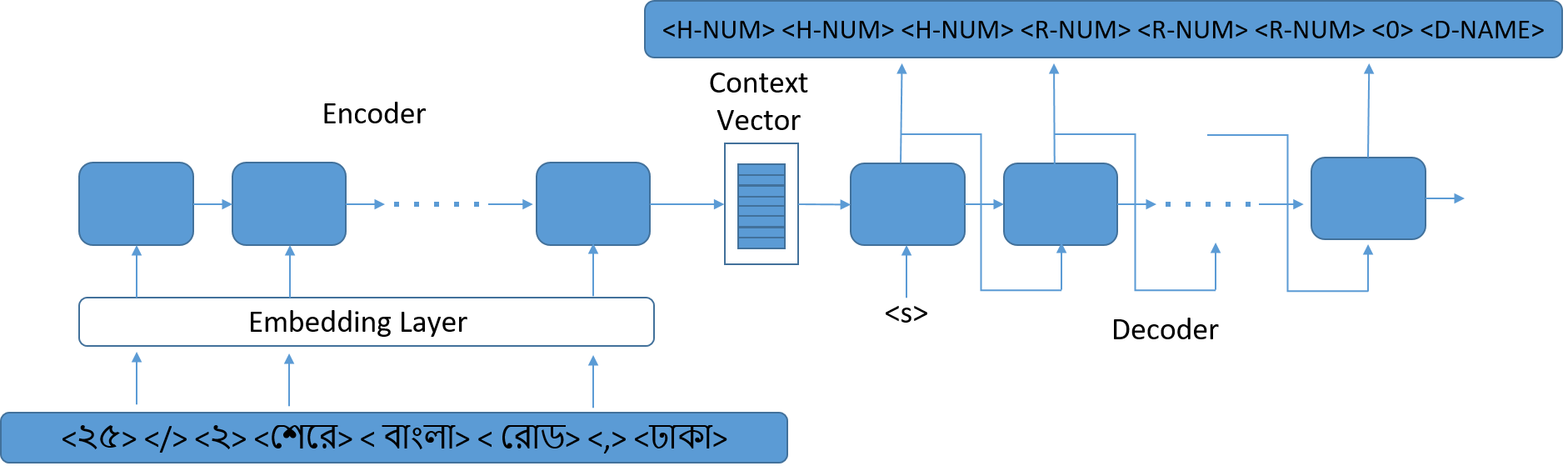}
  \caption{Sequence-to-sequence Encoder-Decoder models using RNN, LSTM, and Bi-LSTM units for the Bangla address text parsing problem}
  \label{fig: Encoder-Decoder Parsing Model Architecture}
\end{figure*}

\begin{figure*}[!h]
  \centering
  \includegraphics[width=0.7\textwidth]{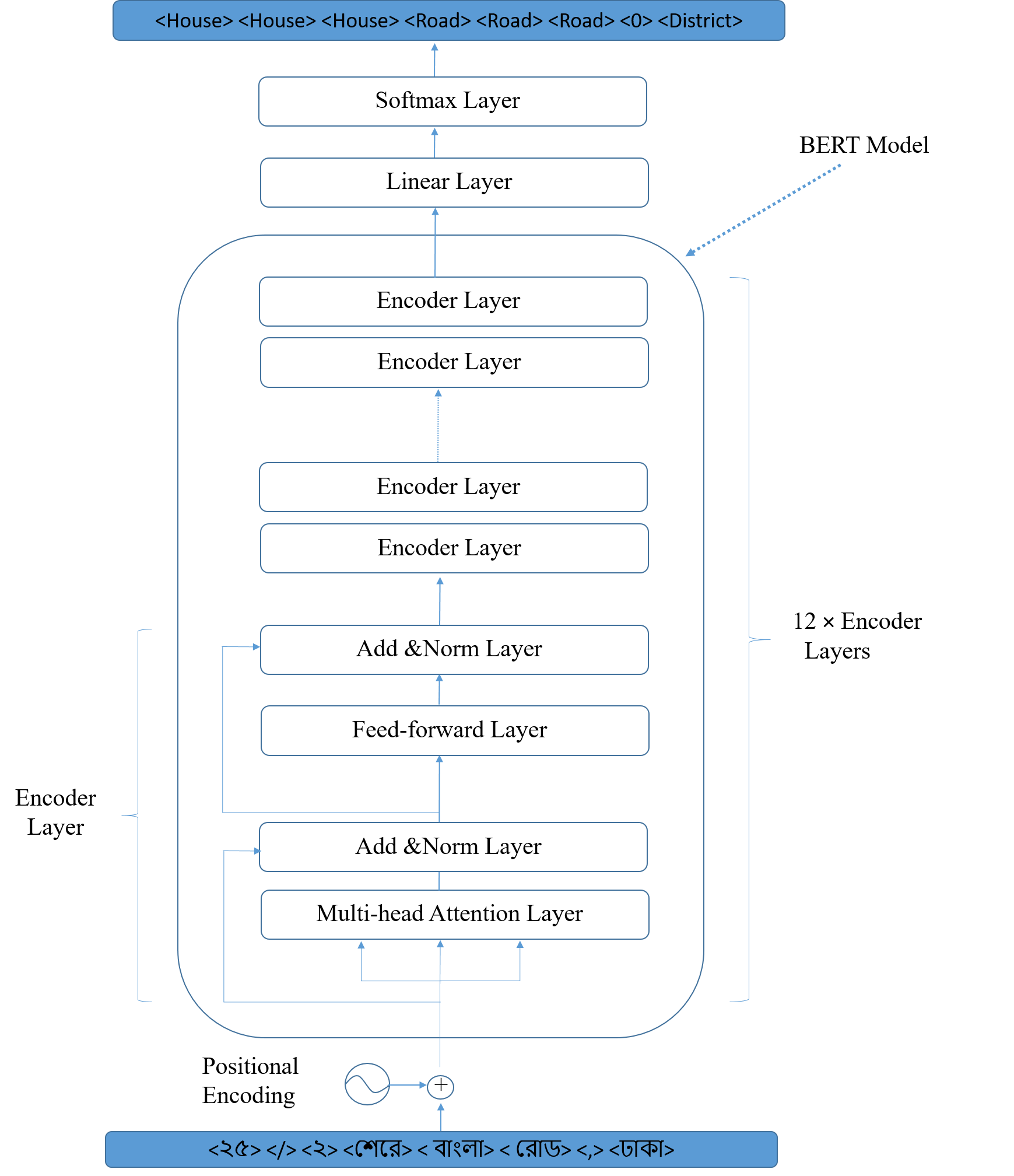}
  \caption{Token classification model for the Bangla address text parsing problem using the transformer-based pre-trained language}
  \label{fig: Transformer-based Parsing Model Architecture}
\end{figure*}

\subsection{Address Text Parser Model}
The next step is to parse the address text into different address components such as house number, road number/name, area name, thana name, and district name. We can consider the address text parsing problem to both sequence-to-sequence modeling and token classification problem. When we consider the address text parsing problem as a sequence-to-sequence modeling problem, then we design the encoder-decoder model using RNN, LSTM, and Bi-LSTM units. On the contrary, when we consider the address parsing problem as a token classification problem, then we design the transformer-based pre-trained language model for Bangla address text parsing. 

The input of the Bangla address text parser model is a sequence of word tokens and the output is a sequence of address components. For sequence-to-sequence modeling problems, the Encoder-Decoder framework is one of the effective model architectures. We have utilized the Encoder-Decoder model architecture for Bangla address text parser model. Figure \ref{fig: Encoder-Decoder Parsing Model Architecture} shows an overview of the Encoder-Decoder model for the Bangla address parser model. In this research work, we train and evaluate different sequence-to-sequence Encoder-Decoder models using RNN, LSTM, and Bi-LSTM units.

\subsubsection{Transformer-based Pre-trained Language Model for Bangla Address Text Parser Model Architecture}
In the address text parsing task, we tag each token of the address text as an address component. Therefore, we can consider address text parsing as a token classification problem. We propose a token classification model for the Bangla address text parsing problem using the transformer-based pre-trained language named Banglabert \cite{bhattacharjee2021banglabert}. Banglabert is a BERT-based pretrained language model in Bangla. Bidirectional Encoder Representations from Transformers (BERT) is designed based on the Encoder network of the original transformer model architecture. Banglabert has been trained on a large corpus of 27.5 GB Bangla text collected from 110 Bangla websites. Banglabert model is a popular pretrained model for downstream tasks like text classification, token classification, and question answering for Bangla language.

We can see an overview of the proposed model architecture in Figure \ref{fig: Transformer-based Parsing Model Architecture}. The Encoder network consists of 12 encoder layers where each encoder layer has a multi-head attention layer followed by add \& norm layer and a feed-forward layer followed by another add \& norm layer. Moreover, there are two residual connections - one residual connection is from the input of the multi-head attention layer to the first add \& norm layer and another residual connection is from the input of the feed-forward layer to the second add \& norm layer. Finally, there is a linear layer and a softmax layer at the end of the model architecture.

In this research work, we train and evaluate the transformer-based pre-trained language model for Bangla address text parsing using the Bangla Address Parsing (Bn-AP) dataset. Moreover, we present a comparative analysis among transformer-based pre-trained language models and sequence-to-sequence Encoder-Decoder models with RNN, LSTM, and Bi-LSTM units for Bangla address text parsing.

\section{Datasets}\label{datasets}
In this section, we present the datasets developed to train proposed model architectures for different sub-problems of the end-to-end system. We discuss the data collection or creation procedure for each of the datasets. Moreover, we show the data preprocessing steps followed before using the dataset for the training and evaluation of different model architectures.

\begin{figure}[!h]
  \centering
  \includegraphics[width=\columnwidth]{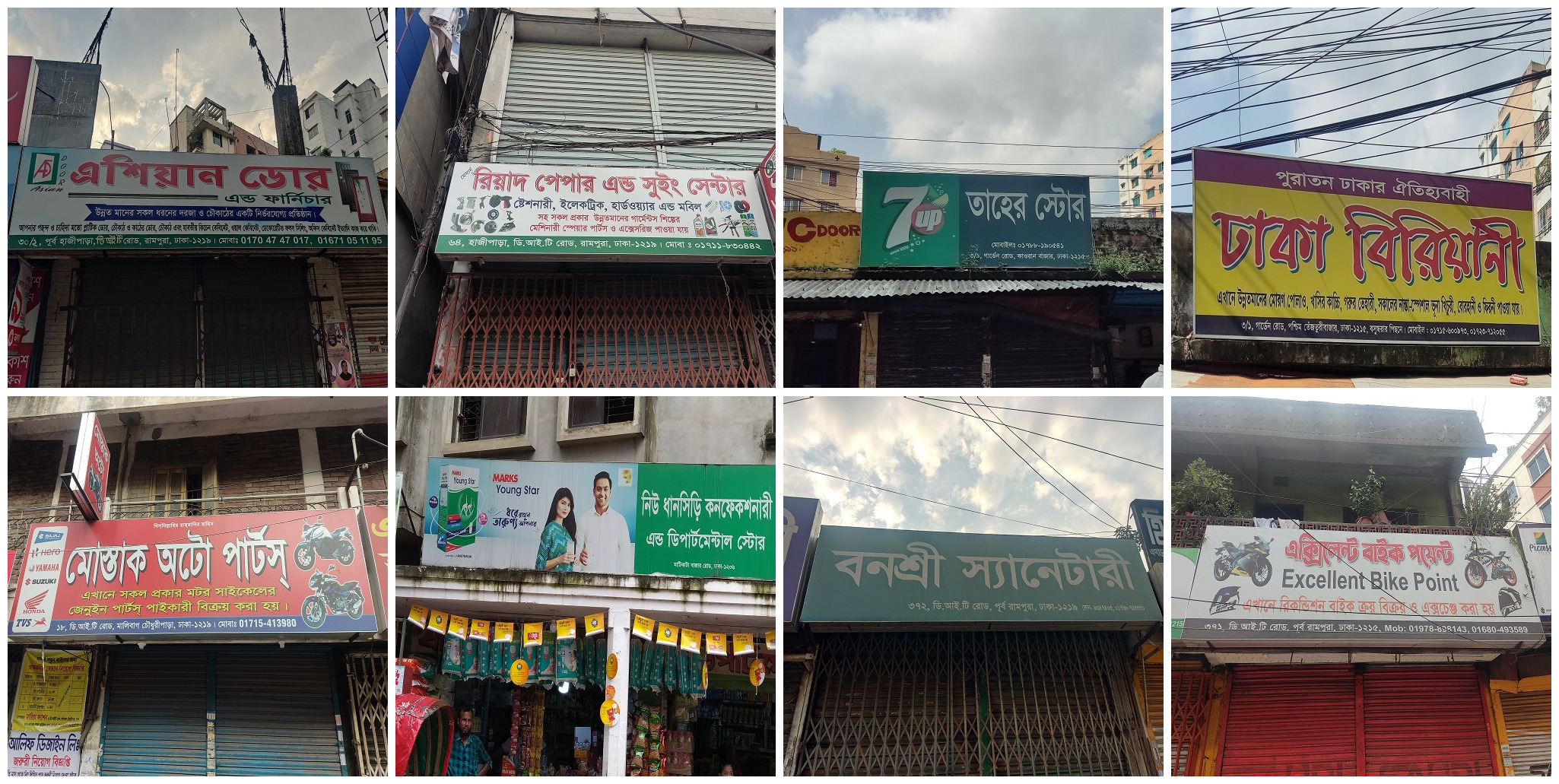}
  \caption{Sample natural scene images from the collected raw images}
  \label{fig: Sample Dataset}
\end{figure}

\subsection{Bangla Signboard Detection (Bn-SD) Dataset}
The signboard detection model detects signboards from the raw image with the background. Recognizing address text from raw images with complex backgrounds is different without detecting the signboard portion first.  However, signboard detection is a challenging problem in a developing country like Bangladesh where no standard signboard design is followed. Moreover, signboards come in various sizes, colors, and orientations, depending on their purpose, location, and local regulations. There is no existing Bangla signboard detection dataset available to train a signboard detection model. Therefore, in the research work, we have created a novel Bangla signboard detection dataset named Bn-SD by collecting more than 16000 natural scene images containing Bangla signboards. Different Yolo-based object detection models have been trained on the Bn-SD dataset.
\begin{figure}[!h]
  \centering
  \includegraphics[width=\columnwidth]{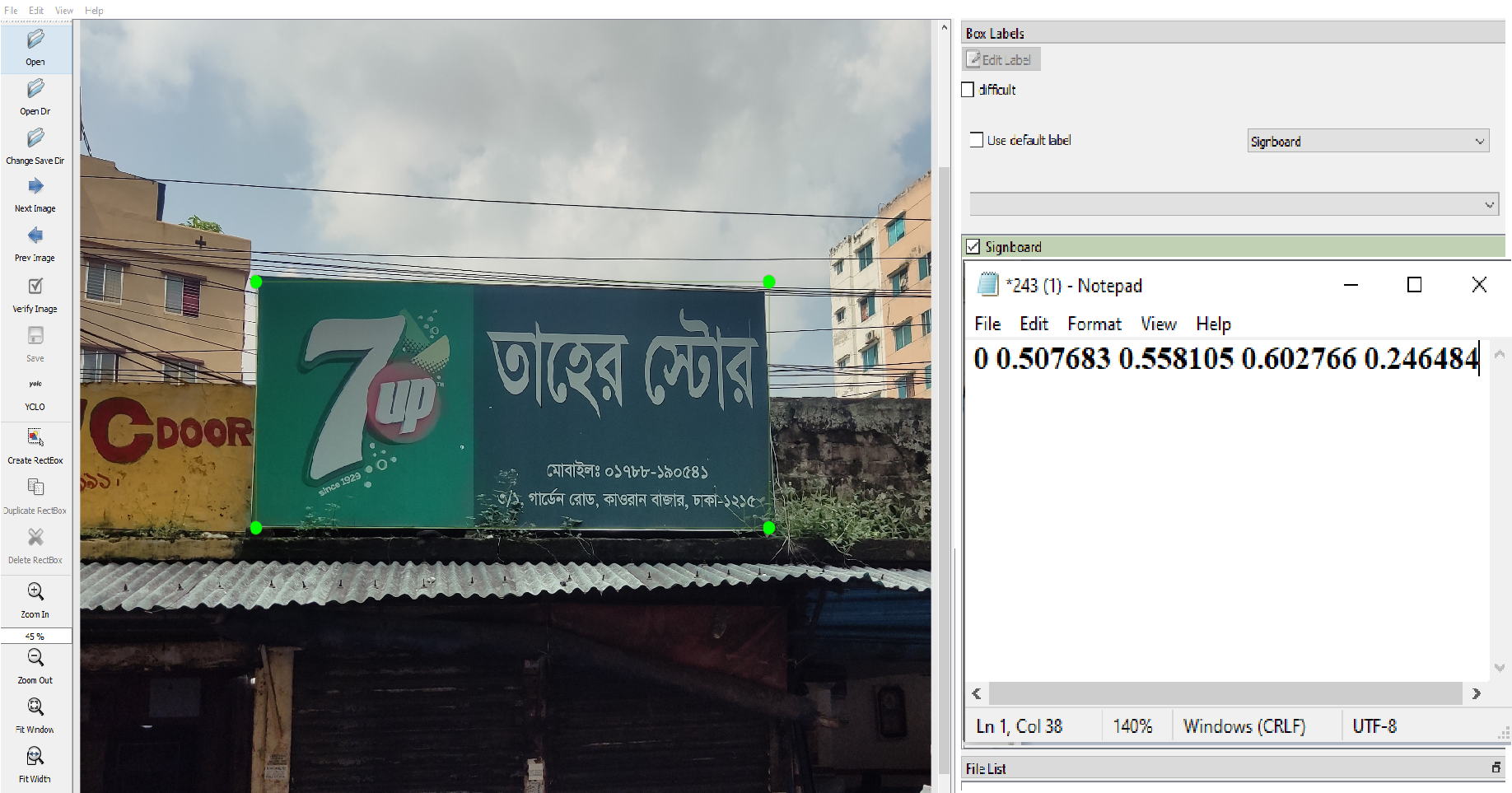}
  \caption{The annotation process of the Bangla Signboard Detection (Bn-SD) Dataset using LabelImg. For each image, the LabelImg tool creates a txt label file containing the class number (0 for only one class named signboard) and 4 values representing the relative location of the rectangular box.}
  \label{fig: signboard label}
\end{figure}

Dhaka is a densely populated area with more than 23 million people and the population is increasing at almost 3.3\% per year \footnote{\url{https://shorturl.at/dejoQ}}. To serve such a large population, there are an increasing number of marketplaces and commercial areas around Dhaka city. Therefore, there are numerous natural scenes containing signboards with Bangla text. We have collected more than 16000 natural scene images containing Bangla signboards. To collect the natural scene images, we have only selected Dhaka city which is the capital of Bangladesh. Figure \ref{fig: Sample Dataset} presents some sample natural scene images from the collected raw images. Before utilizing the collected dataset, we have to annotate the raw images with an appropriate object detection labeling tool. From raw images, we create the Bangla Signboard Detection (Bn-SD) dataset by annotating it with an object detection labeling tool named LabelImg \footnote{https://github.com/heartexlabs/labelImg}. LabelImg annotates the scene image with rectangular boxes around the signboard area. Figure \ref{fig: signboard label} shows an overview of the annotation process. The total size of the Bn-SD dataset is 27.2 GB after annotating using LabelImg.

\subsection{Bangla Address Detection (Bn-AD) Dataset}
After detecting the signboard from the raw image, the next step is to detect the address text portion from the signboard. We frame the address text detection from the signboard as an object detection problem. We introduce a new dataset by cropping the signboard area of the original natural scene images. Different Yolo-based object detection models have been trained on the Bangla Address Detection (Bn-AD) dataset.
\begin{figure}[!h]
  \centering
  \includegraphics[width=\columnwidth]{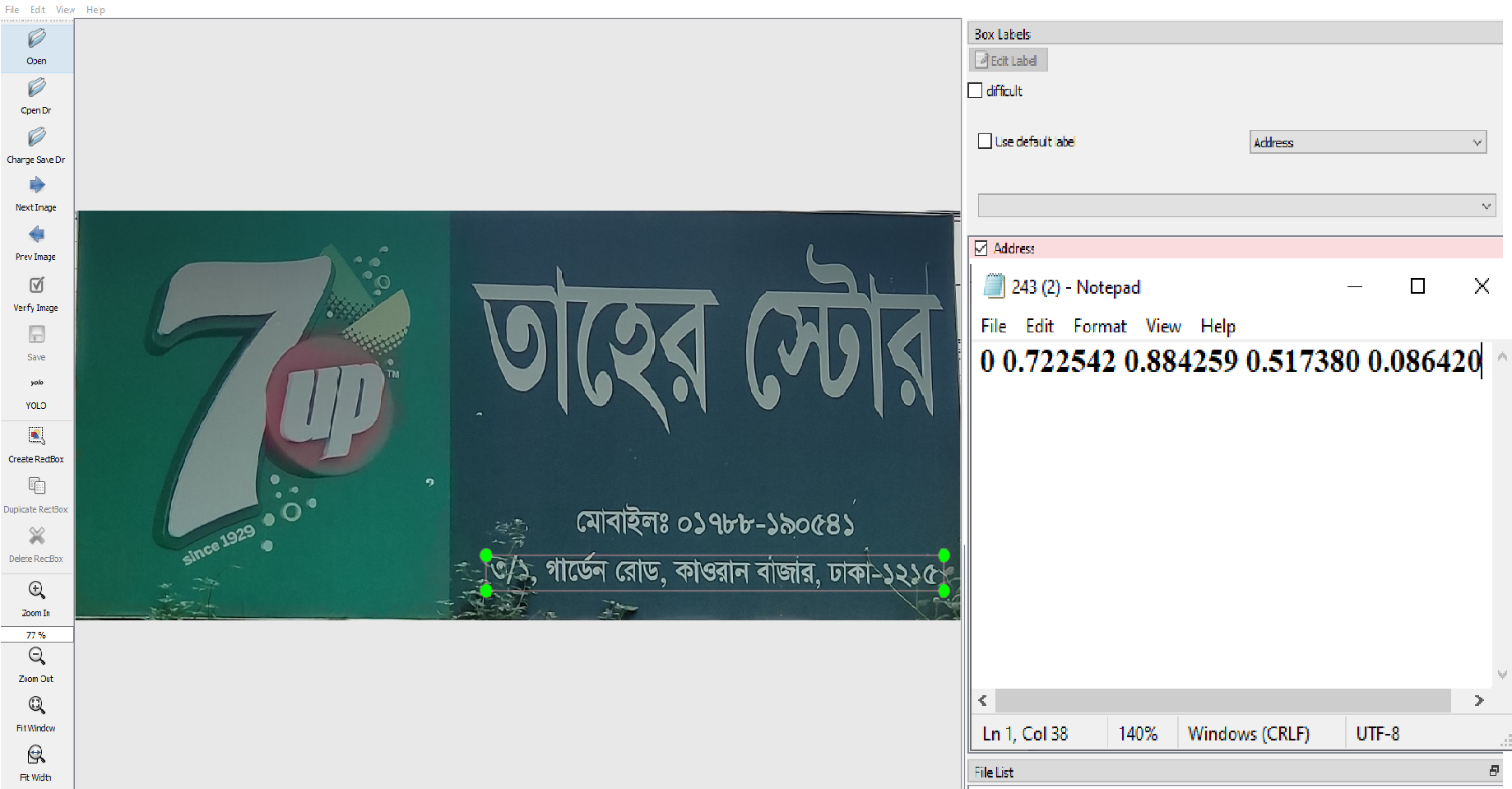}
  \caption{The annotation process of the Bangla Address Detection (Bn-AD) Dataset using LabelImg. For each image, the LabelImg tool creates a txt label file containing the class number (0 for only one class named address) and 4 values representing the relative location of the rectangular box.}
  \label{fig: address label}
\end{figure}

From the Bn-SD dataset, we separate the raw images which contain signboards with address information. There are more than 8000 signboard images with address information. We then create a novel dataset by cropping the signboard area from the separated raw images. We create the Bangla Address Detection (Bn-AD) dataset by annotating rectangular boxes around the address area with the LabelImg object detection labeling tool. Figure \ref{fig: address label} shows an overview of the annotation process for the Bn-AD dataset. The total size of the Bn-AD dataset is 5.01 GB after annotating using LabelImg.

\subsection{Raw Address Text Corpus}
We have collected a large dataset of full address text from different areas of Dhaka city such as Dhanmondi, Hajaribag, Jigatola, Rayer Bazar, Shankar, and Tolarbag. The full address consists of the address segments/components - House number, Road number/name, Area name, Thana name, District name. We use the raw address text corpus to create the datasets for Bangla address text recognition, address text correction, and address text parser models. We apply different data augmentation techniques on the raw address text before using it to create new datasets.




\subsection{Synthetic Bangla OCR (Syn-Bn-OCR) Dataset}
\begin{figure}[!h]
  \centering
  \includegraphics[width=\columnwidth]{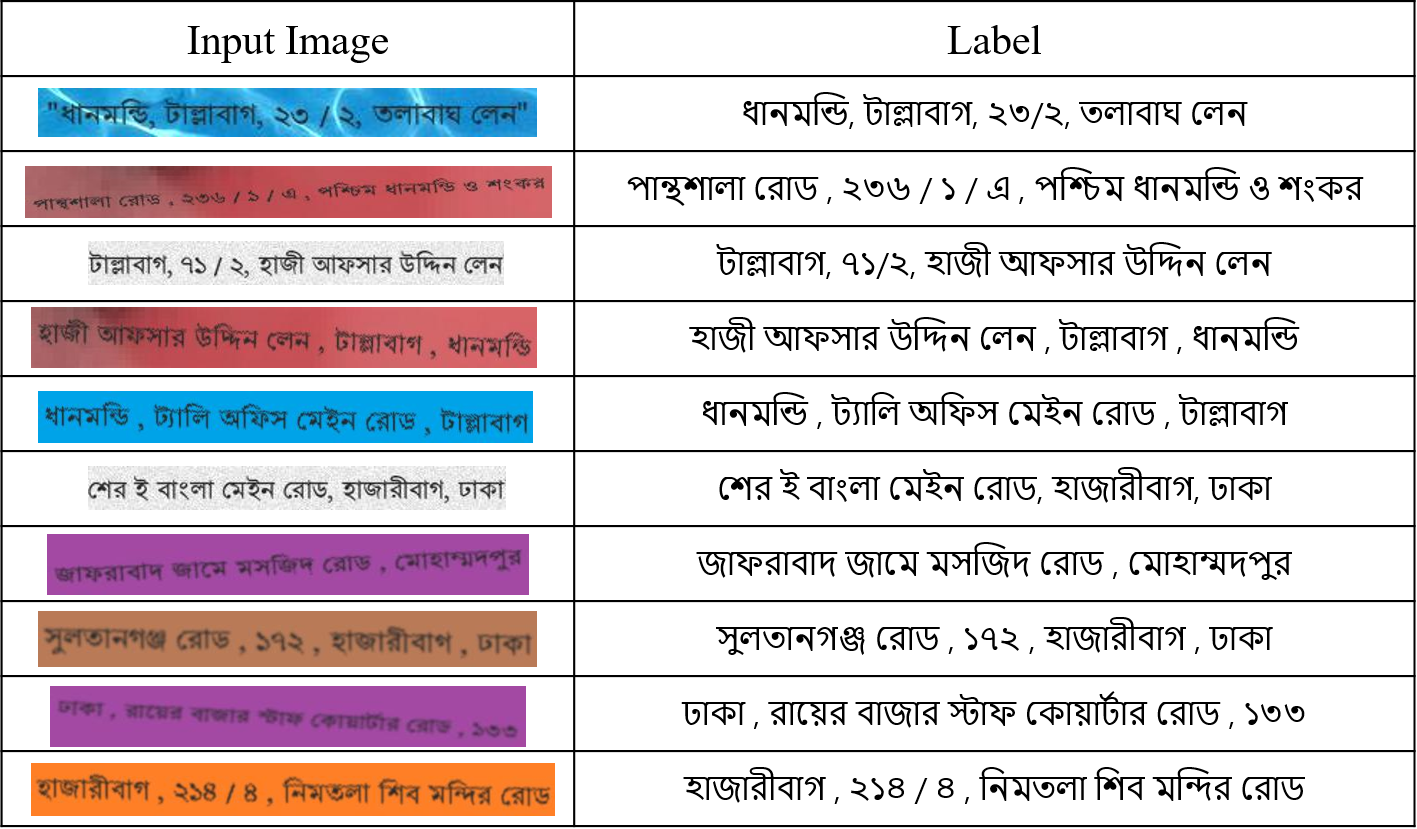}
  \caption{Some sample data from the Synthetic Bangla OCR (Syn-Bn-OCR) dataset generated by the TextRecognitionDataGenerator tool}
  \label{fig: Sample syn-bn-ocr Dataset}
\end{figure}

To train the CTC-based or Encoder-Decoder model for Bangla address text recognition, we need a large label dataset. However, as Bangla is a low-resource language, such a large label dataset is not available. Therefore, we create a synthetic dataset named Synthetic Bangla OCR (Syn-Bn-OCR) dataset by using a text recognition data generator named TextRecognitionDataGenerator \footnote{https://github.com/Belval/TextRecognitionDataGenerator}. Figure \ref{fig: Sample syn-bn-ocr Dataset} shows some sample data from the Syn-Bn-OCR dataset. The TextRecognitionDataGenerator is a popular tool to generate synthetic text recognition datasets where we need to provide only a raw Bangla address text corpus. The TextRecognitionDataGenerator generates images with Bangla address text on them.

\subsection{Synthetic Bangla Address Correction (Syn-Bn-AC) Dataset} \label{subsection: Syn-Bn-AC}
Due to the complexity of the scene text recognition, the address text recognition model recognizes the address text with some errors in the character sequence. We introduce the Bangla address text correction model to improve the performance of the Bangla address text recognition model by post-correction. However, there is no available Bangla address text correction dataset in Bangla. Therefore, we need to develop a Bangla text correction dataset before training and evaluating the Bangla address text correction model. 

\begin{figure}[!h]
  \centering
  \includegraphics[width=\columnwidth]{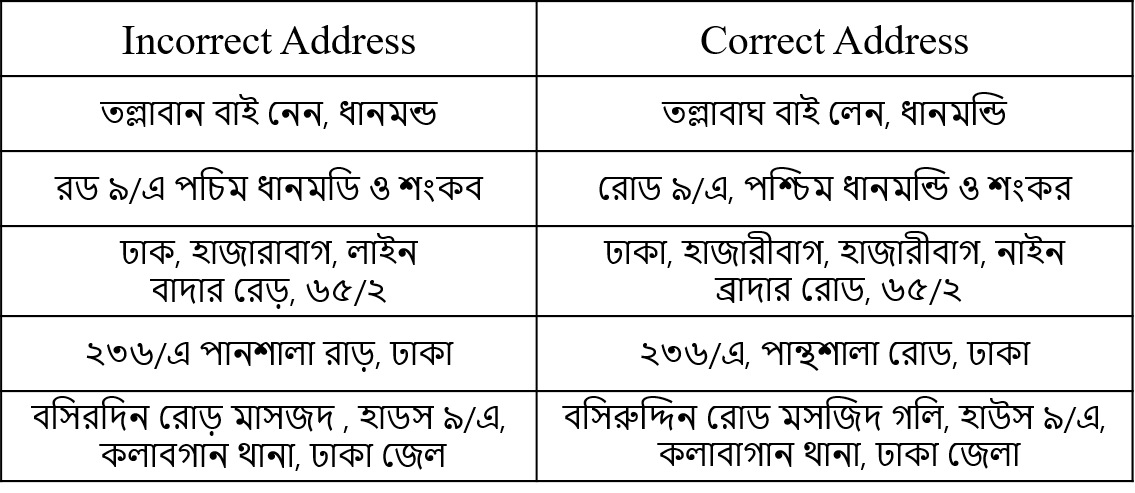}
  \caption{Some samples from the Synthetic Bangla Address Correction (Syn-Bn-AC) dataset}
  \label{fig: Sample bn-ac Dataset}
\end{figure}

 We have created a synthetic Bangla address correction (Syn-Bn-AC) dataset by utilizing the raw address text dataset. We introduce errors in the address text by randomly inserting, replacing, or removing Bangla characters into the address text. The number of modifications depends on the length of the address text. We add randomly 1\% to 10\% errors on the address text depending on the length of the original address text. Figure \ref{fig: Sample bn-ac Dataset} shows some sample pairs of incorrect and correct addresses from the Synthetic Bangla Address Correction (Syn-Bn-AC) dataset. The modified incorrect address text is the input to the address text correction model where the original address text is the output.

\subsection{Bangla Address Parsing (Bn-AP) Dataset}
We have created a novel Bangla Address Parsing (Bn-AP) Dataset by labeling the raw full address text dataset from different areas of Dhaka city such as Dhanmondi, Hajaribag, Jigatola, Rayer Bazar, Shankar, and Tolarbag. During labeling the full address text, we consider 5 different address components such as house number, road number/name, area name, thana name, and district name. We extend the size of the dataset by applying different augmentation techniques: randomly removing address components, randomly swapping two address components, reversing the address text, and 
removing punctuations.

\begin{figure}[!h]
  \centering
  \includegraphics[width=\columnwidth]{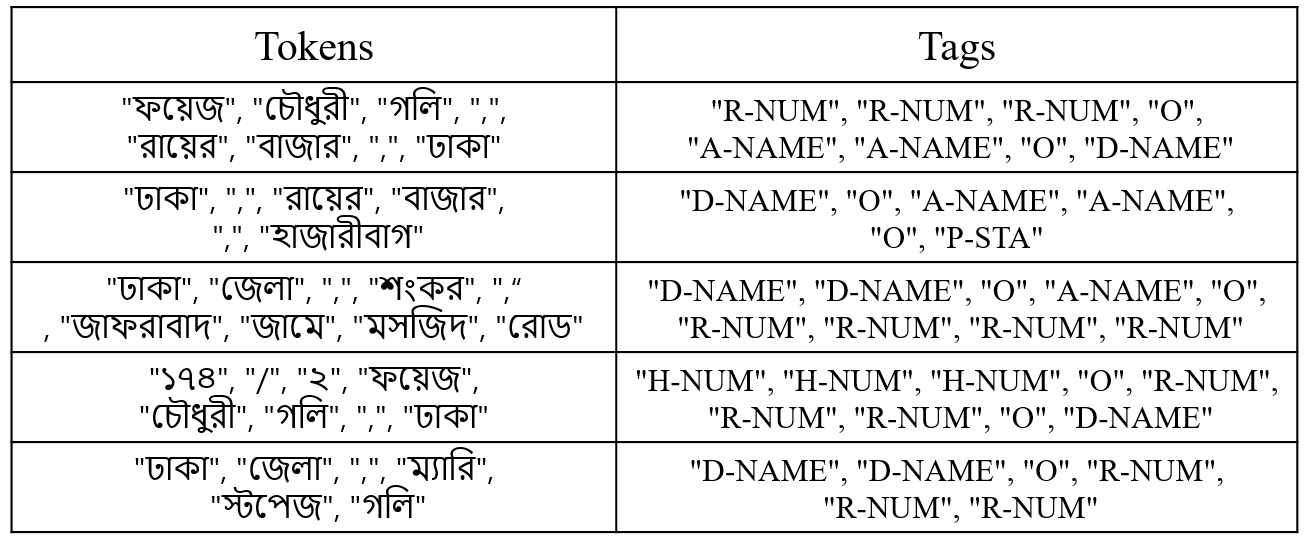}
  \caption{Some sample data from the Bangla Address Parsing (Bn-AP) dataset}
  \label{fig: Sample bn-ap Dataset}
\end{figure}

We create a sequence of tokens from the address text and label each token with a tag. Therefore, each data sample is a pair of a sequence of tokens and a sequence of tags. Figure \ref{fig: Sample bn-ap Dataset} shows some sample data from the Bangla Address Parsing (Bn-AP) dataset.

\section{Experimental Results and Discussions}\label{results}
In this section, we discuss the experimental results and discussion of our research work. In our system pipeline, there are five different deep learning-based models for signboard detection from the raw image, address text detection from the signboard image, Bangla address text recognition from the cropped address portion, address text correction, and finally address text parsing. We have proposed different model architectures for each of the sub-problems. In the following sections, we discuss the experimental results during the training and evaluation of the proposed model architectures. Moreover, we conduct a comparative study among different possible model architectures for each of the sub-problems. 

\subsection{Environment Setting}
The training, validation, and testing of the different proposed model architectures are performed on a computer with hardware configurations: an Intel processor (model number i7-7700k) with 8 MB Intel smart cache, 4 processing cores, and 8 threads running with 4.50GHZ max turbo frequency; an Nvidia GTX 1070 GPU with 8GB of VRAM; a DDR5 RAM with 16 GB operating at a clock speed of 5200 MHz; and storage of 1 TB with 512 GB SSD and 512 GB HDD. Moreover, we need to use the Google Colab platform with hardware configuration: a processor with Intel(R) Xeon(R) CPU @ 2.20GHz; a 16GB NVIDIA Tesla T4 GPU; and a 12.68 GB system RAM; and a 78.2 GB disk storage.

\subsection{Signboard Detection Model}
To develop a signboard detection model, we train and evaluate different recent Yolo-based object detection models such as YOLOv3, YOLOv4, and YOLOv5 using our novel Bn-SD dataset. During training and testing on the model architectures, we consider the following parameter setting and evaluation metrics.

\subsubsection{Parameter Setting}
While training the Yolo-based models, we need to set some parameters to their recommended values. The Yolo-based model takes a fixed-size image as input. Therefore, the spatial dimension of the input image is set to  $416 \times 416$. As we are working with RGB images, the number of channels is 3. To ensure the stability of the gradient during training, we set the momentum to 0.95 and weight\_decay to 0.0005. Burn\_in is assigned to 1000 so that the learning rate steadily increases for the first 1000 batches until it reaches a specified value of 0.001. It is recommended to assign the max\_batches to 6000 for only 1 class detection. The steps parameter is set to (80\% of max\_batches, 90\% of max\_batches) which is equal to (4800, 5400). After 4800 steps, the learning rate is multiplied by a factor of 0.1 so that it would be decreased to 0.0001. After 5400 steps, the learning rate is again multiplied by a factor of 0.1 so that it would be decreased to 0.00001. By decreasing the learning rate at steps 4800 and 5000, we get the nearly perfect model \cite{bochkovskiy2020yolov4}. The number of filters required is calculated by using the following Equation \ref{equation: filter}. As the number of classes is 1, using Equation \ref{equation: filter}, we set the number of filters to 18. We choose a batch size of 64 and a subdivision of 16.
\begin{equation}
    filter = (5 + number\_of\_classes) \times 3
    \label{equation: filter}
\end{equation}

\subsubsection{Evaluation Metrics}
For evaluating different Yolo-based Bangla signboard detection models, we use standard evaluation metrics such as Average Precision (AP). To calculate the AP, it is required to find the Intersection Over Union (IoU), Precision (P), and Recall (R).

IoU is the ratio of the area overlap and the area of union between the predicted and the ground truth bounding box. IoU refers to the accuracy of bounding box prediction and the range of the IoU is 1 to 0. We calculate the value of IoU using the following Equation \ref{equation: IoU}.
\begin{equation}
    IoU = \frac{Area\_of\_Overlap}{Area\_of\_Union}
    \label{equation: IoU}
\end{equation}
According to the PASCAL VOC metrics \cite{everingham2010pascal, padilla2020survey}, we calculate different evaluation metrics such as Precision, Recall, and AP when the IoU is greater than 50\% (IoU $\ge$ 0.5). That means if the IoU is greater than 50\%, then we consider that the signboard is correctly detected. 

The true positive instance refers to a correct detection with $IoU \ge 50\%$, the false positive is defined as an incorrect detection with $IoU < 50\%$, and finally false negative refers to the ground truth signboard is not detected due to the low confidence level. The confidence level indicates how likely a bounding box contains an object and how accurate the boundary box prediction is.  Precision is the measure of the accuracy of positive predictions and can be calculated by the ratio of the number of true positive instances and all truly predicted instances(sum of all true positive and false positive instances). Recall is the ability to correctly identify all positive instances and can be calculated by the ratio of the true positive instances and all true instances in the dataset (sum of all true positive and false negative instances). The formulas to measure the Precision and Recall are shown in Equation \ref{equation: Precision} and \ref{equation: Recall} respectively.
\begin{equation}
    P = \frac{True\_Positive}{True\_Positive + False\_Positive}
    \label{equation: Precision}
\end{equation}
\begin{equation}
    R = \frac{True\_Positive}{True\_Positive + False\_Negative}
    \label{equation: Recall}
\end{equation}

F1 score is an evaluation metric that combines Precision and Recall into a single metric by taking the harmonic mean. We show the formula to calculate the F1 score in Equation \ref{equation: f1_score}.
\begin{equation}
    F1\ Score = \frac{2 \times Precision \times Recall}{Precision + Recall}
    \label{equation: f1_score}
\end{equation}

However, the F1 score represents only single points on the precision-recall curves. Average Precision (AP) overcomes the drawback of F1 score. The AP represents the overall precision-recall curves into a single evaluation metric. Therefore the AP is a more acceptable and comprehensive metric to evaluate the signboard detection models. We calculate the value of AP using the following Equation \ref{equation: ap}. The AP is the weighted average of the precision at n thresholds where the weight is the recall increment for each threshold.
\begin{equation}
    AP = \sum_{k=0}^{k = n -1} [R(k) - R(k-1)] \times P(k)
    \label{equation: ap}
\end{equation}

\subsubsection{Training and Testing Results for Signboard Detection Model}
\begin{figure}[!tb]
  \centering
  \includegraphics[width=\columnwidth]{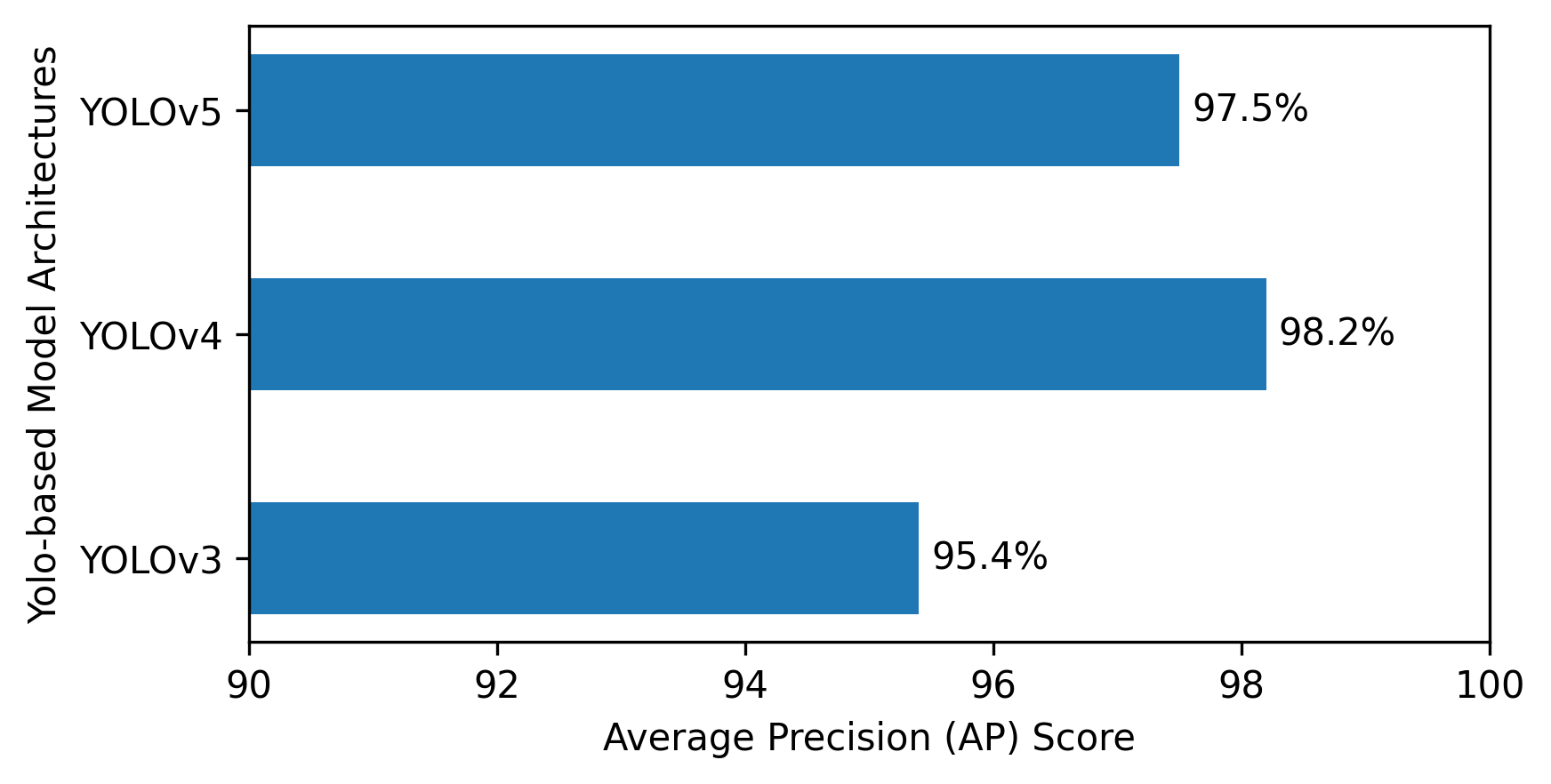}
  \caption{AP scores of different Yolo-based model architectures for signboard detection model}
  \label{fig: ap-score-signboard-detection-model}
\end{figure}

Before training the Bangla signboard detection model, we have divided the Bn-SD dataset into train (80\%), validation (10\%), and test (10\%) sets. By utilizing the above parameter setting and evaluation metrics, we have trained the different Yolo-based models such as YOLOv3, YOLOv4, and YOLOv5 for Bangla signboard detection using our novel Bn-SD dataset. We have trained each model for 6000 iterations. We have conducted a comparative study among YOLOv3, YOLOv4, and YOLOv5 models for Bangla signboard detection by calculating the AP score on the test dataset.  After evaluating the models on the test dataset, the YOLOv3 model shows an AP score of 95.4\%, whereas the YOLOv4 and YOLOv5 models provide AP scores of  98.2\% and 97.5\% respectively. Figure \ref{fig: ap-score-signboard-detection-model} shows the AP scores of different Yolo-based model architectures in a bar chart for the signboard detection model.

The YOLOv4 and YOLOv5 both utilize CSPDarknet53 as the backbone model. The main difference of YOLOv5 is that it utilizes the Pytorch Python library during implementation. Therefore, the YOLOv4 and YOLOv5 models provide nearly the same result due to the underlying similarity of the model architectures.

\subsection{Address Text Detection Model}
The address text detection model detects the address text portion from the input signboard image. The training process of the address text detection model is similar to the signboard detection model. To develop an address text detection model, we train and evaluate different recent Yolo-based object detection models such as YOLOv3, YOLOv4, and YOLOv5 using our novel Bn-AD dataset. During training and testing on the model architectures, we consider the same parameter setting and evaluation metrics applied for the signboard detection model. 

\subsubsection{Training and Testing Results for Address Text Detection Model}
\begin{figure}[!h]
    \centering
    \includegraphics[width=\columnwidth]{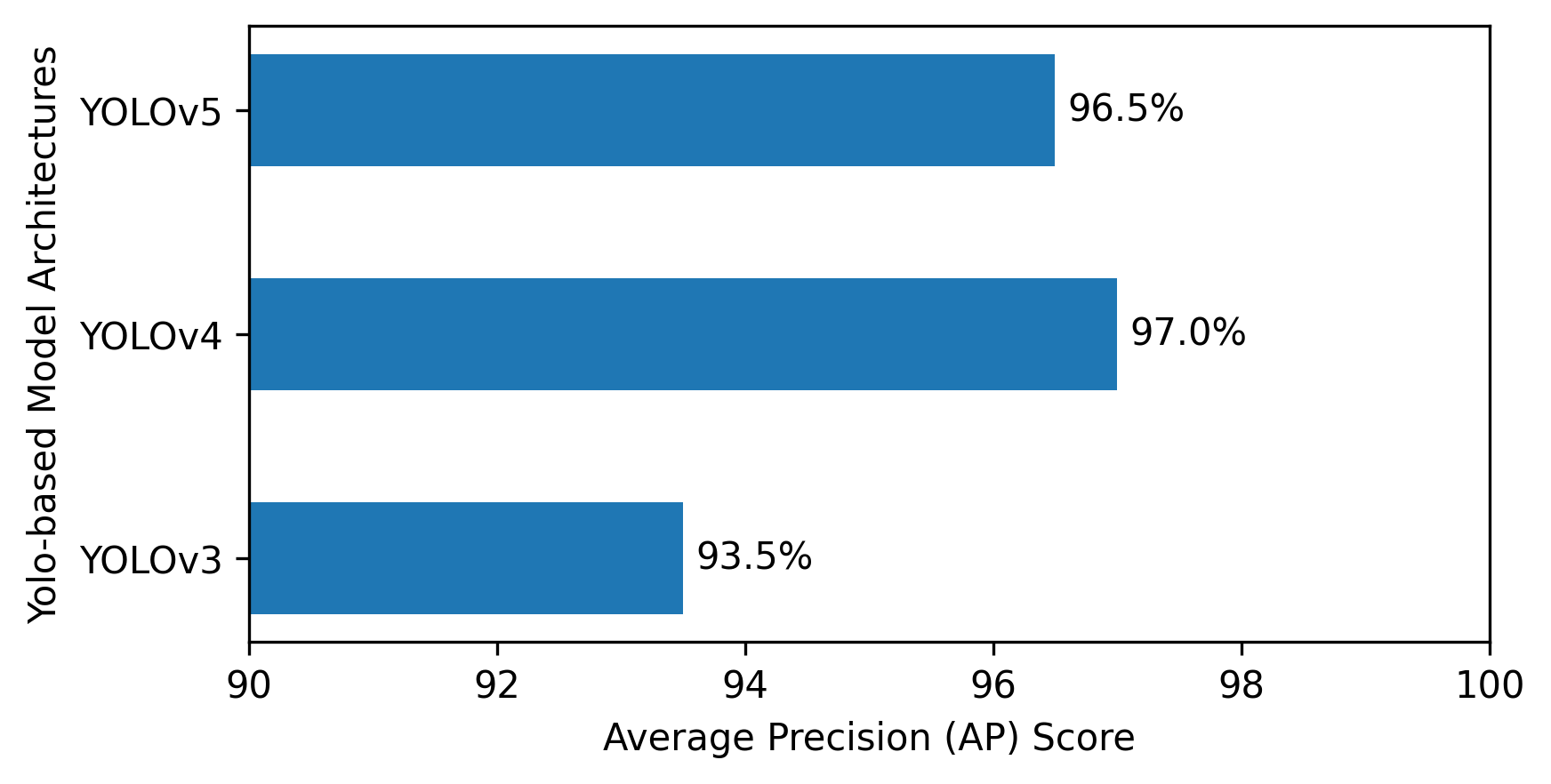}
    \caption{AP scores of different Yolo-based model architectures for address text detection model}
    \label{fig: ap-score-address-detection-model}
\end{figure}

We have partitioned the Bn-AD into three subsets: train (80\%), validation (10\%), and test (10\%) set. We have trained different Yolo-based models such as YOLOv3, YOLOv4, and YOLOv5 using the novel Bn-AD dataset for Bangla address text detection. We have trained each model for 10000 iterations. After the training process, we have conducted a comparative study among YOLOv3, YOLOv4, and YOLOv5 models by calculating the AP score for address text detection.  We have evaluated the AP score on the test dataset for each Yolo-based model. The YOLOv3 model provides an AP score of 93.5\%, whereas the YOLOv4 and YOLOv5 models show AP scores of 97.0\% and 96.5\% respectively for address text detection. Figure \ref{fig: ap-score-address-detection-model} shows the AP scores of different Yolo-based model architectures in a bar chart for the address text detection model. The YOLOv4 and YOLOv5 models provide nearly similar results due to the similarity of the backbone model of CSPDarknet53. 

\begin{table*}[h]
\centering
\begin{tabular}{|c|c|c|}
\hline
\textbf{Type of Framework} &\textbf{Model Architecture} & \textbf{WRA} \\ \hline
\multirow{4}{*}{CTC-based} & VGG+Bi-LSTM+CTC (Baseline) & 91.5\% \\ \cline{2-3} 
& RCNN+Bi-LSTM+CTC        & 92.9\%                   \\ \cline{2-3} 
& GRCL+Bi-LSTM+CTC         & 92.2\%                   \\ \cline{2-3} 
& ResNet+Bi-LSTM+CTC       & \textbf{94.5\%}                   \\ \hline
\multirow{4}{*}{Encoder-Decoder} & VGG+Bi-LSTM+Attention (Baseline) & 91.4\% \\ \cline{2-3} 
& RCNN+Bi-LSTM+Attention       & 92.2\%                   \\ \cline{2-3} 
& GRCL+Bi-LSTM+Attention              & 92.8\%                   \\ \cline{2-3} 
& ResNet+Bi-LSTM+Attention       & 93.5\%                   \\ \hline
\end{tabular}
\caption{A comparison among different Bangla address text recognition models by calculating word recognition accuracy on the test set}
\label{table: wra for text recognition}
\end{table*}

\subsection{Bangla Address Text Recognition Model}
After detecting the address text portion from the signboard image, the next step is to recognize the Bangla address text from it. To design a Bangla address text recognition model, we propose both CTC-based and Encoder-Decoder model architecture. In the case of CTC-based model architectures, we use different pretrained feature extractors such as VGG, RCNN, GRCL, and ResNet, recurrent layers with  Bi-LSTM units, and transcription layers with CTC loss. The Encoder-Decoder network capitalizes the attention mechanism for designing the Bangla address text recognition model. To train the different Bangla address text recognition model architectures, we consider the following parameter setting and evaluation metrics.

\subsubsection{Parameter Setting}
We select the shape of the input image as $64 \times 600$. During the training loop, the batch size is set to 32. We choose the Adadelta optimizer, which is a modified version of the Adagrad optimizer. We set some hyper-parameters of the Adadelta optimizer such as the initial learning rate to 0.01, $\beta_1$ to 0.9, $\rho$ to 0.95, $\epsilon$ to 0.00000001, and the grad\_clip to 5.

\subsubsection{Evaluation Metrics}
We use Word Recognition Accuracy (WRA) as the evaluation metric to evaluate the Bangla address text recognition model. WRA is the ratio between the number of correctly recognized words ($W_r$) and the overall count of ground-truth words ($W$). We calculate the WRA using the following Equation \ref{equation: WRA}.
\begin{equation}
    WRA = \frac{W_r}{W}
    \label{equation: WRA}
\end{equation}

\subsubsection{Training and Testing Results for Bangla Address Text Recognition Model}
We have created Synthetic Bangla OCR (Syn-Bn-OCR) dataset with 100k labeled images using a popular synthetic data generator named TextRecognitionDataGenerator. We have partitioned the Syn-Bn-OCR dataset into three subsets: train set (80\%), validation set (10\%), and test set (10\%).

Using the stated parameter setting and evaluation metrics, we have trained different CTC-based model architectures (VGG+Bi-LSTM+CTC, RCNN+Bi-LSTM+CTC, ResNet+Bi-LSTM+CTC, and GRCL+Bi-LSTM+CTC) and different Encoder-Decoder model architectures (VGG+Bi-LSTM+Attention, RCNN+Bi-LSTM+Attention, ResNet+Bi-LSTM+Attention, and GRCL+Bi-LSTM+Attention) using the Syn-Bn-OCR dataset. We can consider the VGG+Bi-LSTM+CTC and VGG+Bi-LSTM+Attention models as the baseline model as VGG is the simpler feature extractor compared to RCNN and ResNet. We have trained each model architecture before convergence. Table \ref{table: wra for text recognition} shows the word recognition accuracy for each different Bangla address text recognition model on the test dataset. We have found that the ResNet+Bi-LSTM+CTC model shows the best result with a word recognition accuracy of 94.5\%. We have better results for the CTC-based model as the CTC-based layer has less dependency on the language models to capture the final character sequence \cite{long2021scene}. However, the results of the attention-based model differ from the CTC-based with only a very small margin.

\begin{table*}[h]
\centering
\begin{tabular}{|c|c|c|c|c|}
\hline
\textbf{Model Architecture} & \textbf{Precision} & \textbf{Recall} & \textbf{F1 Score} & \textbf{Accuracy}\\ \hline
Encoder-Decoder model with RNN (Baseline) & 0.89 & 0.90 & 0.895 &  90.16\%   \\ \hline
Encoder-Decoder model with LSTM & 0.92 & 0.94 & 0.93 &  93.30\%   \\ \hline
Encoder-Decoder model with Bi-LSTM & 0.93 & 0.95 & 0.94 & 94.55\%   \\ \hline
Token classification model with Banglabert  & \textbf{0.96} & \textbf{0.97} & \textbf{0.965} &  \textbf{97.32\%}   \\ \hline
\end{tabular}
\caption{A comparison among different Bangla address text parser models on the test set}
\label{table: address parsing}
\end{table*}

\subsection{Address Text Correction Model}
We propose an address text correction model to improve the performance of the Bangla address text recognition model by post-correction. The address text correction model automatically corrects the incorrect output character sequence of the Bangla text. Sequence-to-sequence Encoder-Decoder model is the most effective model to generate the correct character sequence from the incorrect character sequence using contextual information. We have designed a transformer-based encoder-decoder model for Bangla address text correction. To train the address text correction model architecture, we consider the following parameter setting and evaluation metrics.

\subsubsection{Parameter Setting}
Using the training phase, we select the batch size of 32 and the initial learning rate of 0.00005 and the number of warmup steps of 10000, and the weight decay rate of 0.01. For the tokenization step, we choose the Byte-Pair Encoding (BPE) tokenizer \cite{sennrich2015neural}.

\subsubsection{Evaluation Metrics}
During the training and testing phase, we evaluate the address text correction model using the Word Level Accuracy (WLA) with is similar to the evaluation metric used for the Bangla address text recognition model.  If the number of correct words in the output sequence is $W_c$ and the number of words in the ground-truth sequence is $W$, then we calculate the WLA using the following Equation \ref{equation: WLA}.
\begin{equation}
    WLA = \frac{W_c}{W}
    \label{equation: WLA}
\end{equation}

\subsubsection{Training and Testing Results for Address Text Correction Model}
We have created the synthetic Bangla address correction (Syn-Bangla-AC) dataset with 60k pairs of incorrect and correct address text. We have partitioned the Syn-Bangla-AC dataset into three sub-sets: train set (80\%), validation set (10\%), and test set (10\%). By considering the above parameter setting and evaluation metrics, we have trained the sequence-to-sequence transformer-based Encoder-Decoder model using the Syn-Bangla-AC dataset for Bangla text address correction. We have trained the model before convergence. We have evaluated the model on the test dataset and found a word-level accuracy of 98.1\%.

\subsection{Address Text Parser Model}
We have proposed two different types of address text parser models by considering the parsing as sequence-to-sequence modeling and token classification problem. We design sequence-to-sequence encoder-decoder models with RNN, LSTM, and Bi-LSTM units to parse the address text. Moreover, we propose a token classification model for Bangla address text parsing problem using the transformer-based pre-trained language named banglabert. While training the model architectures, we consider the following parameter setting and evaluation metrics.

\subsubsection{Parameter Setting}
While training the Bangla address text parser model, we choose a batch size of 32, an initial learning rate of 0.0001, a number of warmup steps of 5000, and a weight decay rate of 0.01. For the tokenization step, we choose the WordPiece tokenization algorithm \cite{bhattacharjee2021banglabert}. 

\subsubsection{Evaluation Metrics}
We evaluate the Bangla address text parser model during the training and testing phase using Accuracy, Precision, Recall, and F1 Score as the evaluation metrics. To calculate the evaluation metrics, we use a popular sequence labeling evaluator Python library named seqeval \footnote{https://github.com/chakki-works/seqeval}. Sample calculations of the evaluation metrics are found in the HuggingFace token classification pipeline \footnote{https://shorturl.at/adprM}.

\subsubsection{Training and Testing Results for Address Text Parser Model}
To extend the novel Bangla Address Parsing (Bn-AP) dataset, we apply different augmentation techniques such as randomly removing address components, randomly swapping two address components, revising the address, and removing punctuation. The final augmented Bn-AP dataset contains 30k labeled data. We have divided the Bn-AP dataset into three subsets: training set (80\%), validation set (10\%), and testing set (10\%).

We have trained the sequence-to-sequence Encoder-Decoder model with RNN, LSTM, and Bi-LSTM units  and finetuned the token classification model using Banglabert model before convergence. The Encoder-Decoder model with RNN units is the simplest model and is considered the baseline model for the address text parsing problem.  Table \ref{table: address parsing} shows the Accuracy, Precision, Recall, and F1 Scores for different Bangla address text parser models on the test dataset. We have found that the token classification model using the Banglabert model provides the best results with a Precision of 0.96, Recall of 0.97, F1 Score of 0.965, and Accuracy of 97.32\%.

\section{Conclusion and Future Works}\label{conclusion}

In this research work, we have developed an end-to-end system for detecting, recognizing, and parsing the Bangla address text from natural scene images containing signboards using deep learning-based approaches. We have developed manually annotated or synthetic datasets for detecting, recognizing, correcting, and parsing address text from the natural scene. We have trained and evaluated different Yolo-based model architectures for detecting signboards from natural scene images and the address text from signboard images. We have conducted a performance analysis among different CTC-based and Encoder-Decoder models with attention mechanisms for Bangla address text recognition and found the best-performing model architecture. We have introduced a novel address correction model using a sequence-to-sequence transformer network to improve the performance of Bangla text recognition by post-correction. Finally, We have developed a Bangla address parser using the state-of-the-art transformer-based pre-trained language model.

In the future, we will extend our novel signboard dataset for multi-modal analysis by utilizing both the visual context and the textual context found in the signboard images. By creating a synthetic Bangla OCR dataset using a general text corpus, we can train a general Bangla scene text recognition model that can recognize Bangla text from natural scene images. We can explore the possibility of using the transform-based model for the sequence modeling layer in the text recognition model. Finally, we can train a general text correction model by training the sequence-to-sequence transformer-based model architecture with a Bangla text correction dataset created by using a general text corpus.

\bibliography{anthology,custom}

\end{document}